\documentclass{article} 
\usepackage[dvipsnames]{xcolor}
\usepackage{collas2025_conference,times}
\usepackage{subfig}
\usepackage{easyReview}
\usepackage{booktabs}
\usepackage{colortbl}

\usepackage{multicol}
\usepackage{multirow}

\usepackage{soul}
\usepackage{xcolor}
\definecolor{lightorange}{RGB}{255, 220, 180}
\usepackage[normalem]{ulem}  

\usepackage{amsmath,amsfonts,bm}

\def\eqref#1{equation~\ref{#1}}

\def\1{\bm{1}}

\DeclareMathAlphabet{\mathsfit}{\encodingdefault}{\sfdefault}{m}{sl}
\SetMathAlphabet{\mathsfit}{bold}{\encodingdefault}{\sfdefault}{bx}{n}

\usepackage{hyperref}
\hypersetup{
    colorlinks=true,
    linkcolor=red,
    filecolor=magenta,
    urlcolor=blue,
    citecolor=purple,
    pdftitle={Overleaf Example},
    pdfpagemode=FullScreen,
    }

\title{Beyond Cosine Decay: On the effectiveness of Infinite Learning Rate Schedule for Continual Pre-training}

\author{\parbox{\textwidth}{\centering
\vspace{0.75cm}
    Vaibhav Singh$^{1,2*}$  \hspace{-10pt}
    \qquad Paul Janson$^{1,2}$\thanks{Equal contribution: (vaibhav.singh,paul.janson)@mila.quebec} \hspace{-10pt}
    \qquad Paria Mehrbod$^{1,2}$  \hspace{-10pt}\\
    \qquad Adam Ibrahim
    \qquad Irina Rish$^{1, 3}$
    \qquad Eugene Belilovsky$^{1, 2}$ \vspace{5pt}
    \qquad Benjamin Thérien$^{1, 3}$ \hspace{-10pt}\\
    \textnormal{{ $^1$Mila -- Quebec AI Institute; $^2$Concordia University, Montréal; $^3$Université de Montréal }}
}}

\makeatletter
\def\blfootnote{\xdef\@thefnmark{}\@footnotetext}
\makeatother

\collasfinalcopy

\begin{document}

\maketitle
\begin{abstract}
The ever-growing availability of unlabeled data presents both opportunities and challenges for training artificial intelligence systems. While self-supervised learning (SSL) has emerged as a powerful paradigm for extracting meaningful representations from vast amounts of unlabeled data, existing methods still struggle to adapt to the non-stationary, non-IID nature of real-world data streams without forgetting previously learned knowledge. Recent works have adopted a repeated cosine annealing schedule for large-scale continual pre-training; however, these schedules (1) inherently cause forgetting during the re-warming phase and (2) have not been systematically compared to existing continual SSL methods.
In this work, we systematically compare the widely used cosine schedule with the recently proposed infinite learning rate schedule and empirically find the latter to be a more effective alternative. Our extensive empirical evaluation across diverse image and language datasets demonstrates that the infinite learning rate schedule serves as a flexible and robust alternative without being restricted to a fixed iteration budget. For instance, in a small-scale Masked AutoEncoder (MAE) pre-training setup, it performs competitively against several strong baselines from the literature. We then scale up our experiments to larger MAE pre-training and autoregressive language model pre-training. Our results show that the infinite learning rate schedule continues to show stable and effective behavior—maintaining knowledge retention and adaptability across tasks, while eliminating the need to know the total training iterations in order to fix a termination point. Code is available at  \href{https://github.com/Pauljanson002/beyond-cosine}{https://github.com/Pauljanson002/beyond-cosine.}
\end{abstract}

\section{Introduction}
Self-supervised~\citep{balestriero2023avi} pre-training has emerged as a transformative paradigm in machine learning~\citep{he_masked_2022, radford2018improving, devlin2018bert}, catalyzing the development of foundational models in vision \citep{clip, dinov2, segment_anything, shang2024theia} and language \citep{foundationmodels, Achiam2023GPT4TR, touvron2023llama, zhao2023survey} that are now widely deployed across diverse applications~\citep{chatgpt, guo2025deepseek,Anthropic}. These models are known for their massive parameter counts and extensive training on vast amounts of data, often developing impressive general-purpose capabilities unexpectedly during pre-training~\citep{brown2020language, wei2022emergent}. 

While foundation models have demonstrated remarkable success on static tasks, adapting them to evolving data—such as the continuous influx of new textual information \citep{soldaini2024dolma, dclm, german, code} and the emergence of novel visual concepts \citep{prabhu2023categories, seo2024just}—remains a major challenge. This is primarily due to the high costs of retraining and the risk of catastrophic forgetting \citep{mccloskey1989catastrophic} induced by significant distributional shifts. While recent studies \citep{ke2023continual, qiao2024learn, yildiz2024investigating, reuse_nvidia} provide guidelines for continual pre-training in language modeling, systematic approaches that seamlessly integrate into existing language model pre-training pipelines remain lacking. In the context of computer vision, conventional CL approaches such as regularization techniques~\citep{kirkpatrick2017overcoming, Li17learning, aljundi2018memory}, and architectural modifications~\citep{douillard2022dytox, yan2021dynamically}— struggle to scale effectively to modern foundation models. These challenges stem from two core limitations: (1) their inability to generalize to self-supervised learning objectives and large-scale datasets, and (2) the architectural constraints they impose, which may not align with the diverse model architectures in contemporary use.

\begin{figure}[t!]
  \centering
  \includegraphics[trim={70 20 80 30},clip,width=0.95\textwidth]{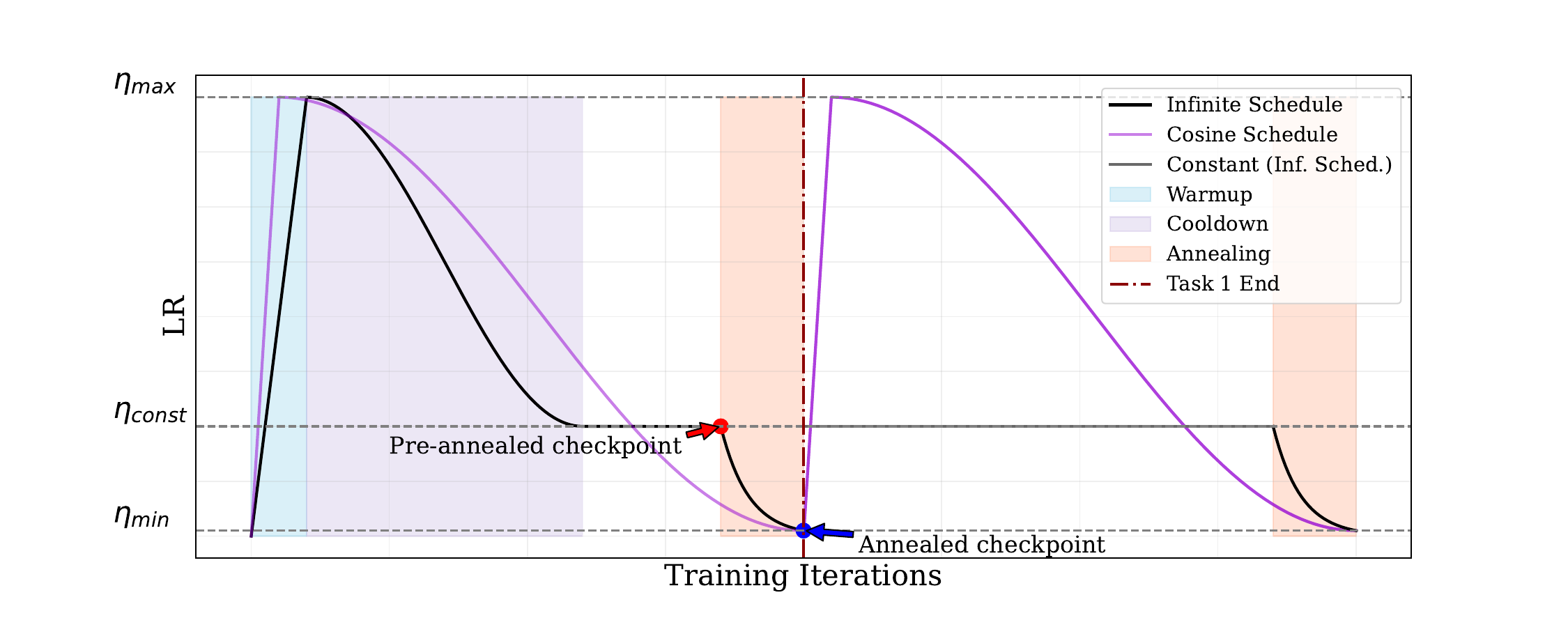}
\caption{\small \textbf{Comparing an Infinite Learning Rate Schedule with Repeated Cosine Annealing for Two-Task CL.} We illustrate the key differences between the \textbf{infinite learning rate schedule} and the \textcolor{Plum}{\textbf{cosine schedule}}. The infinite schedule consists of four distinct phases: warmup, cooldown, constant, and annealing (see legend). The \textcolor{Brown}{\textbf{vertical line}} indicates the completion of Task 1. For continual training, infinite scheduler offers two strategic checkpointing options: the \textcolor{red}{\textbf{pre-annealed checkpoint}} at learning rate $\eta_{const}$ before annealing begins, which enables training continuation on new tasks, and the \textcolor{blue}{\textbf{annealed checkpoint}} at $\eta_{min}$ after annealing completes for deployment. In contrast, the cosine schedule lacks the constant phase, making it less flexible for CL scenarios.
  }
  
  \vspace{-5mm}
  \label{fig:cosine_inf_schedule}
\end{figure}

Most approaches for continually pre-training foundation models typically utilize a repeated cosine annealing schedule~\citep{loschilov2017warmrestarts} with fixed duration \citep{gupta2023continual,  defazio2023optimal, ibrahim2024simple, reuse_nvidia, stability_gap}. In this schedule, the learning rate undergoes an initial warmup phase to reach its maximum value, followed by a cosine decay that gradually reduces it to the minimum value precisely at the end of training (\textcolor{Plum}{purple} in \autoref{fig:cosine_inf_schedule}). This implicitly assumes a terminal point in the training process, which severely limits the future pre-training on new datasets without undergoing significant forgetting. This fundamental limitation inhibits true continuous adaptation, as traditional learning rate schedules inevitably decay to near-zero values, effectively preventing further meaningful updates to the model. Secondly, re-warming the learning rate from its minimum value causes instability and exacerbates forgetting \citep{ibrahim2024simple}. To overcome this constraint, recent works have explored more flexible \textit{infinite learning rate} schedules that accommodate varying training durations~\citep{zhai_scaling_2022, road_less_scheduled, hu2024minicpm, shen2024power, gele2024scaling}. These schedules consist of four distinct phases: an initial warmup, followed by a decay phase (e.g., cosine, inverse square root) that reduces the learning rate to a constant value, a plateau phase where this rate is maintained, and a final rapid annealing phase near the end of training (\textbf{black} line in \autoref{fig:cosine_inf_schedule}). While these innovations emerged primarily from data-scaling research, their applications have begun to extend into CL, as demonstrated in \citep{garg2023tic, ibrahim2024simple}.

However, these works fail to answer a critical open question: \textit{How do these scheduling approaches behave under distribution shifts, i.e. non-IID data distributions}\footnote{Some previous works exploring infinite LR schedules~\citep{ibrahim2024simple, garg2023tic} considered different datasets stemming from splitting a single original dataset, leading to substantially weaker shifts than those considered in this work.}\textit{?} This scenario is particularly relevant for practical applications where models must continuously adapt to data from diverse domains. For instance, consider the challenge of continually pre-training an English language model to incorporate German. In such scenarios, catastrophic forgetting severely impacts model performance.

In this work, we answer this question by comprehensively analyzing the importance of learning rate schedules for self-supervised continual pre-training. Through extensive experiments across vision and language modalities, we believe that, to the best of our knowledge, we are the first to conduct a detailed comparison of infinite learning rate schedules with repeated annealing. Our results show that infinite schedules offer a flexible and stable approach to mitigating catastrophic forgetting in the non-IID setting, maintaining model performance across diverse data distributions, and offering a competitive advantage in retaining the previous task performance as compared to the repeated cosine with warmup.

This work makes several key contributions:
\begin{itemize}
    \item We present the first systematic study on the impact of learning rate schedules in non-IID self-supervised Continual Learning across both vision and language modalities. 
    
    \item We demonstrate that just using infinite learning rate schedules alone and combined with experience replay can outperform a number of sophisticated continual baselines in the context of self-supervised continual learning (Sec \ref{subsec-cifar10})

    \item We further demonstrate that, across multiple sequential large-scale vision and language pre-training tasks, infinite learning rate schedules either match or outperform repeated cosine annealing, while offering greater flexibility by not requiring a predefined number of data(Sec \ref{MAE_fn}, \ref{llm_results}).




      \item Our results show that the \textbf{Infinite Cosine Schedule} is a flexible and robust alternative to repeated cosine decay for continually pre-training foundation models, demonstrating that it can improve knowledge retention and adaptability relative to the repeated cosine schedule, in challenging non-IID self-supervised learning scenarios across both vision and language modalities.

\end{itemize}

\section{Related Work}
\label{sec:related_work}

\textbf{Continual pre-training (CPT) of Vision Foundation Models}
Continually pre-training Vision Transformers (ViTs) \citep{dosovitskiy2020image, bao2021beit} aims to adapt them to sequential data while mitigating catastrophic forgetting. \citet{wang2022lvt} introduced the Lifelong Vision Transformer (LVT), incorporating inter-task attention to preserve critical weights across tasks. \citet{ye2024task} proposed a task-free dynamic sparse ViT for scenarios without explicit task boundaries. The rise of large-scale foundation models has reshaped CL, particularly Vision-Language Models (VLMs) \citep{clip,janson2022simple,garg2023tic,zhang2024overcoming,singh2024controlling}, where CPT offers an efficient alternative to full retraining. Unlike parameter-efficient adaptation methods \citep{wang2022learning,wang2022dualprompt,smith2023coda}, our work focuses on adapting the whole model. 
 
\textbf{Continual pre-training (CPT) of Large Language Models (LLMs)} Recent studies \citep{scialom_fine-tuned_2022, winata2023overcoming, mehta2023role, gupta2023continual} have outlined strategies for CPT, that learns general representations for diverse downstream tasks. A key theme is the inherent ability of LLMs to accumulate and retain knowledge across tasks \citep{brown2020language}. \citet{cossu2022continualpretraining} demonstrated that CPT mitigates catastrophic forgetting, with self-supervised approaches outperforming supervised ones. Larger pretrained models also exhibit reduced forgetting compared to those trained from scratch, attributed to their increasingly orthogonal class representations \citep{ramasesh2022scale,mirzadeh2022wide}. Additionally, \citet{scialom_fine-tuned_2022} provide evidence that self-supervised pre-training naturally enables CL.

\textbf{Alternatives to Cosine Schedule}
The cosine decay schedule \citep{loschilov2017warmrestarts} is widely used in vision tasks, where stepwise or cyclic learning rates help to escape suboptimal minima during multi-epoch training \citep{no_decay}. For language models, the cosine annealing schedule with a single cycle is the standard \citep{gupta2023continual, reuse_nvidia}, but its reliance on a fixed training step count makes it unsuitable for continuous training. To address this, the Warmup-Stable-Decay (WSD) scheduler \citep{hu2024minicpm} was introduced, enabling continuous training. \citet{shen2024power} further refined this approach, discovering a power law relationship in optimal learning rate patterns, leading to the power scheduler, which applies warmup followed by exponential decay based on token count. \citet{gele2024scaling} challenged the cosine annealing schedules' fixed-duration requirement, proposing constant learning rates with cooldown periods instead. While these advancements enable training without a terminal point, they primarily tackle the problem of data scaling rather than distribution shifts.

In the context of temporal distribution shifts, \citet{garg2023tic} made notable progress by implementing a repeated cosine annealing schedule for continual supervised pre-training of CLIP models, with warmup applied exclusively to the initial task. Their investigation, which included a variant of the WSD scheduler, revealed an important insight: \textbf{cosine rewarming tends to diminish final model performance, leading to their recommendation to utilize checkpoints from the constant phase}. While this finding aligns with \citet{ibrahim2024simple}'s work on CPT of language models, the latter's experiments regarding infinite learning rate schedules focused on in-distribution learning without considering distribution shifts. Our work advances this line of research by extending it in two critical directions
first, by examining LLM pre-training under explicit distribution shift scenarios where catastrophic forgetting will be severe, and second, by expanding the framework to vision foundation models through masked image modeling approaches~\citep{he_masked_2022,fang2023eva}.

\vspace{-2mm}
\section{The Need for Infinite Learning Rate Scheduling: Why It Matters?}
\vspace{-2mm}
\label{sec:infinite_schedule}
Cosine Scheduling has been the de facto learning rate scheduling method for training large-scale models. Within a single cycle, it effectively balances between utilizing high learning rates for rapid optimization and gradually reducing rates to stabilize convergence, with the cycle duration expanding proportionally to the number of training iterations. However, this approach requires knowing the specific number of iterations in advance, which inhibits the ability to continually train a model on new incoming data. Several works~\citep{zhai2022scalingvit, gele2024scaling, ibrahim2024simple, hu2024minicpm} suggest alternatives such as infinite learning rate schedules and warmup-stable-decay(WSD), which offer the flexibility of training without predetermined step counts. These methods inherently support CL by default. An additional advantage of these alternate schedules is the ability to anneal or rapidly decrease the learning rate \textit{near the end} of the training phase. This steep reduction has been shown to dramatically decrease the loss~\citep{schaipp2025surprising}, enabling practitioners to preserve model checkpoints just before the decay phase for subsequent continual training cycles. Further infinite schedules enable targeted replay strategies, where replay buffer configurations can be adjusted dynamically (e.g., during annealing) to emphasize retention of past knowledge, providing more control than fixed-schedule approaches. 
This approach is particularly valuable in scenarios where the best-performing model must be deployed to users while anticipating the acquisition of additional high-quality data in the future, as it enables continued training without a fixed iteration budget, supports agile checkpointing for seamless updates, and allows replay strategies to be tuned dynamically for improved knowledge retention across tasks.

In this work, we investigate the effectiveness of the infinite cosine schedule \cite{ibrahim2024simple} relative to repeated cosine for the CPT of models under strong distribution shifts. We perform a comprehensive comparison of the infinite and cosine schedules across diverse self-supervised learning tasks in both vision and language domains. Through extensive experimentation, we demonstrate that infinite learning rate scheduling not only enhances robustness to distribution shifts but also serves as a better alternative to cosine scheduling by eliminating the need to predefine the training duration.

We define Infinite Cosine Schedule as given in \citet{ibrahim2024simple}: 

\begin{equation}
  \begin{array}{c}
      \text{Inf Cosine}(n) = 
      \begin{cases}
          \frac{n}{N_{w}} \cdot \eta_{max}, & \text{if } n < N_{w} \\[10pt]
          \eta_\textit{const} + \frac{\eta_\textit{max} - \eta_\textit{const}}{2} \cdot \left(1 + \cos \left(\pi\frac{n-N_{w}}{N_{c}-N_{w}}\right)\right), & \text{if } N_{\text{w}} < n \leq N_{\text{c}} \\[10pt]

          \eta_{const}, & \text{if } N_{\text{c}} < n \leq N_{\text{d}} \\[10pt]

          \eta_\textit{const} \cdot \left(\frac{\eta_\textit{min}}{\eta_\textit{const}}\right)^{\frac{n - N_{d}}{t_{a} + N_{d}}} & \text{if } n > N_{d}
      \end{cases}
  \end{array}
\end{equation}

where $n$ is the current training step, $\eta_{max}$ and $\eta_{min}$ denote the maximum and minimum learning rates respectively, and $N_{w}, N_{c}$ $N_{d}$ denote number of warmup steps, cooldown steps, and decay steps respectively, each specifying the transition points between the phases. $t_a$ denotes the amount of annealing steps required to achieve a converged checkpoint. 
As illustrated in \autoref{fig:cosine_inf_schedule}, the infinite learning rate schedule progresses through four distinct phases: during the warmup phase (until $N_{w}$), the learning rate increases linearly from $0$ to $\eta_{max}$. It then transitions to a cosine cooldown schedule until reaching $N_{c}$, followed by a stable constant phase until $N_{d}$, before finally annealing exponentially until the end of the task. We additionally define the cooldown proportion as $P=N_{c}/N$. For continually pre-training on subsequent tasks, we can efficiently resume training from the pre-annealed checkpoint at the constant learning rate $\eta_{const}$.  Note that the subsequent tasks will only consist of constant phase and annealing, eliminating the need for rewarming.

\section{Experimental setup}
\label{section:experiments}

Our experiments span both vision and language  domains focusing on significant distribution shifts across a sequence of datasets $\mathcal{D}_0, \mathcal{D}_1, \ldots, \mathcal{D}_{N-1}$. We first evaluate infinite schedule on a small-scale MAE pre-training \citep{he_masked_2022}, comparing it to CL baselines (Sec~\ref{sec:mae-cpt}). Next, we scale up to large-scale vision datasets with significant distribution shifts (Sec \ref{sec:multi-dataset-vision}). Finally, we demonstrate its generalizability by continually pre-training LLMs across diverse distributions (Sec \ref{section:cl_setup_llm}).
 
\subsection{Continual pre-training of MAEs}
\label{sec:mae-cpt}
We use Masked Autoencoders (MAE) 
\citep{he_masked_2022} for vision pre-training, leveraging their alignment with language models and strong performance in masked image modeling \citep{fang2023eva,singh2023effectiveness}. As described by \citet{he_masked_2022}, MAE pre-training masks a subset of image patches and reconstructs the original image using a Vision Transformer (ViT) \citep{dosovitskiy2020image} encoder-decoder architecture. After pre-training, the decoder is discarded, and the encoder serves as a feature extractor for downstream vision tasks. Additional details regarding MAE pre-training are provided in Appendix \ref{appn:mae-pretrain}.

To validate our hypothesis on infinite learning rate schedules, we conduct an experiment with a small-scale MAE CPT on CIFAR-10~\citep{cifar10}, using a controlled setting for rigorous baseline evaluation. The dataset is divided into five sequential tasks, each with two classes introduced in label order (0-9). We employ a ViT-tiny~\citep{dosovitskiy2020image} to match the scale of CIFAR-10, with our implementation based on~\citet{mae_cifar}. We use a lightweight decoder with learned positional embeddings to reconstruct the masked patches. We train for 400 epochs with a batch size of 512. Hyperparameters for this small scale experiment are provided in Appendix \ref{appn:hyp-vision-tiny}.

\textbf{Baselines and Adaptations:}  
We compare our approach with the following CL baselines, adapting them for self-supervised pre-training: \textbf{Sequential Fine-tuning}: trains sequentially without mitigating forgetting, serving as the primary baseline. \textbf{Experience Replay (ER)}~\citep{rolnick2019experience}: maintains a memory buffer with \{40\%, 50\%\} samples of prior tasks, sampled uniformly. Each batch contains equal proportion of current task data and randomly sampled data from replay buffer. \textbf{Memory Aware Synapses (MAS)}~\citep{aljundi2018memory}: estimates parameter importance by measuring how changes affect the model output, then penalizes updates to important weights. We adapted it for self-supervised learning by computing importance of weights from the L2 norm of the encoder's output, with a regularization $\lambda = 0.75$. \textbf{Learning without Forgetting (LwF)}~\citep{Li17learning}: preserves knowledge by distilling responses from the previous model version. We modified it for self-supervised learning with feature distillation on the encoder's output, weighted by $\alpha = 0.75$. \textbf{GDumb}~\citep{prabhu2020gdumb}: Uses stratified sampling to maintain a balanced buffer. The model resets to random initialization for each new task and trains from scratch on buffer data. For evaluation, we use standard CL metrics from \citet{lopez2017gradient}: Average Accuracy (Acc), Forward Transfer (FWT), and Backward Transfer (BWT), defined in Appendix \ref{appn:eval-metrics}.

\subsection{Large-scale MAE Pre-training across multiple distributions}
\label{sec:multi-dataset-vision}
\textbf{Datasets:} Our pre-training pipeline utilizes three carefully selected large-scale datasets ($N=3$). The CPT sequence begins with ImageNet~\citep{russakovsky_imagenet_2015} ($\mathcal{D}_0$), having 1.28M object-centric images across 1,000 categories, providing a foundation in object recognition. Next, Places2 subset~\citep{zhou2017places} ($\mathcal{D}_1$) introduces a distribution shift with 1M scene-understanding images spanning 365 categories. Finally, FireRisk~\citep{shen2023firerisk} ($\mathcal{D}_3$) presents a substantial shift to remote sensing with 91K satellite images for environmental monitoring. This progression increases distribution shifts, transitioning from object recognition to scene understanding followed by aerial imagery. 

\textbf{Evaluation:} Our evaluation strategy measures both task-specific performance and cross-task knowledge transfer using linear probing. After pre-training on each dataset $\mathcal{D}_i$, we freeze the encoder $f_{\theta}$ as a fixed feature extractor and train a linear classifier $h_{\psi_i}: \mathbb{R}^d \rightarrow \mathbb{R}^{c_i}$ for each task, where $c_i$ is the number of classes. The classifier is optimized with cross-entropy loss, 
and evaluated on task-specific validation sets using classification accuracy, following the same metrics as in Sec.~\ref{sec:mae-cpt}.

\textbf{Implementation:} We build on the PyTorch~\citep{paszke2019pytorch} MAE framework with a ViT-B/16 backbone. For the infinite schedule, we keep a constant learning rate $\eta_{const}=3.75e-5$, while the baseline follows a standard cosine decay schedule with SOTA hyperparameters~\citep{he_masked_2022}. Experiments are conducted with and without a replay buffer of size $B = 0.05 \times |\mathcal{D}_i|$ per task. All models are trained for 300 epochs per task using AdamW~\citep{adamw} with a batch size of 4096. Further implementation and hyperparameter details are given in Appendix \ref{appn:mae-pretrain-large-scale}.
\subsection{Continually pre-training LLMs}
\label{section:cl_setup_llm}
\textbf{Language Datasets:} We consider three datasets for continually pre-training LLMs: DCLM-Baseline \citep{dclm} ($\mathcal{D}_0$), Stack \citep{code} ($\mathcal{D}_1$) and German \citep{german} ($\mathcal{D}_2$). DCLM is a large-scale dataset of natural language text, Stack is a specialized dataset of programming code snippets, and German is a subset of the multilingual OSCAR corpus \citep{german}. The Stack and German datasets were chosen to represent strong, but realistic distribution shifts that are both representative of current CPT applications~\citep{deepseekcoderv2} and allow us to evaluate the model's ability to adapt to new tasks under challenging distribution shifts. We use the standard training splits for both datasets, treating each dataset as locally IID. 

All the three datasets are tokenized through LLaMA-3 tokenizer \citep{llama3tok} owing to its large vocabulary size of 128K tokens (100K from \textit{titktoken}\footnote{\url{https://github.com/openai/tiktoken/tree/main}} and 28K additional tokens for non-English languages). We sample a small subset of 100B tokens from each of the DCLM-Baseline (total = 3T), Stack (total = 744B), and OSCAR (total = 168B) datasets for our CPT experiments. We would like to emphasize that as the domain shifts farther away from the tokenizer's training corpus, the tokenizer might become the key bottleneck to performance. Such scenarios would be unrealistic without a way to adapt the tokenizer. With this in mind, we were careful to select challenging new domains that are still well represented in the tokenizer's vocabulary. Though we did not perform a formal tokenizer coverage analysis, our use of German alongside English datasets aligns with LLaMA-3’s multilingual capabilities \cite{llama3tok}. Stable validation loss across domains indicates no significant tokenizer-data mismatch in practice. We leave the treatment, continual tokenizer adaptation to future work. 

\textbf{Implementation details:} We compare Infinite Cosine Schedule with the de-facto Cosine + Warmup Schedule. We fix $\eta_{max}=3e-4$ and $\eta_{min}=3e-5$ as described in \citep{ibrahim2024simple} for both schedules while varying the cooldown proportion ($N_{warmup}<n \leq N_{const}$) and the $\eta_{const}$ for the infinite schedule. We utilize LLaMA-3 architecture \citep{llama3tok} with 570M parameters, training it as an autoregressive decoder-only transformer with a causal language modeling objective. We use a batch size of $1024$ and sequence length $2048$. Further details on hyperparameters are provided in the \autoref{appn:impl-language}.

\section{Results}
\label{sec:results}

\subsection{Results for pre-training MAE on CIFAR10 }
\label{subsec-cifar10}

\begin{table}[h!]
    \centering
    \renewcommand{\arraystretch}{1.2}
    \resizebox{\textwidth}{!}{
    \begin{tabular}{c|cc|cc|cc|cc|cc|cc@{}}
    \toprule

    \multirow{2}{*}{\textbf{Replay}} 
    & \multicolumn{2}{c|}{\textbf{FT-seq}} 
    & \multicolumn{2}{c|}{\textbf{MAS}} 
    & \multicolumn{2}{c|}{\textbf{LwF}} 
    & \multicolumn{2}{c|}{\textbf{ER}} 
    & \multicolumn{2}{c|}{\textbf{GDumb}} 
    & \multicolumn{2}{c}{\textbf{Ours (Inf Cos)}} \\  
    \cmidrule(lr){2-3} \cmidrule(lr){4-5} \cmidrule(lr){6-7} 
    \cmidrule(lr){8-9} \cmidrule(lr){10-11} \cmidrule(lr){12-13}
     & Acc $\uparrow$ & BWT $\uparrow$ & Acc $\uparrow$ & BWT $\uparrow$ & Acc $\uparrow$ & BWT $\uparrow$ 
     & Acc $\uparrow$ & BWT $\uparrow$ & Acc $\uparrow$ & BWT $\uparrow$ & Acc $\uparrow$ & BWT $\uparrow$ \\ \midrule
    0\%  & 58.16 & -17.65 & 50.44 & -19.11 & 50.52 & -19.78 & - & - & - & - & \textbf{60.03} & \textbf{-12.61} \\  
    40\% & - & - & 50.36 & -18.90 & - & - & 53.98 & -21.55 & 48.76 & -19.51 & \textbf{61.45} & \textbf{-12.76} \\  
    50\% & - & - & 50.91 & -18.37 & - & - & 57.94 & -18.53 & 48.46 & -18.76 & \textbf{62.16} & \textbf{-12.61} \\ \bottomrule
    \end{tabular}
    }
    \caption{\small Average linear probe accuracy (Acc) and Backward Transfer (BWT) (where $\uparrow$ indicates that higher is better) for comparing CL baselines utilizing cosine schedule with Infinite Schedule on CIFAR10 with varying replay (ER) strategies. It can be observed that the infinite schedule (Inf Cos) consistently achieves superior performance compared to the cosine schedule across all experimental configurations.}
    \label{tab:cifar10-vision}
\end{table}
    
\autoref{tab:cifar10-vision} demonstrates that the infinite cosine schedule outperforms the standard cosine, achieving higher average linear probe accuracy and BWT across all tasks in small-scale CPT on CIFAR-10. Specifically, in CPT without experience replay (ER), it improves average accuracy by \textbf{1.87\%} and BWT by approximately \textbf{4\%} over Finetuning (FT-seq) with a repeated cosine schedule.

Interestingly, in this setup, the combination of the repeated cosine schedule and experience replay (ER) degrades model performance, as seen in the comparison between FT-seq and ER with 40\% replay. This decline likely stems from limited data diversity in small datasets, leading the more aggressive re-warming of the repeated cosine schedule to overfitting to the replay buffer. In contrast, the infinite learning rate schedule eliminates rewarming, effectively circumventing these issues. We would like to emphasize that the unexpectedly poor performance of these methods relative to FT-seq stems from a fundamental difference in the continual learning setting. While these baselines were originally designed for incremental supervised classification, our work centers on incremental masked image modeling (MIM) — a self-supervised task with distinct objectives and evaluation protocols. To the best of our knowledge, this is the first systematic evaluation of continual learning baselines in a self-supervised MIM setting, leaving us without alternatives specifically designed for this novel paradigm. Therefore, the results should be interpreted with this context in mind.

While replay behavior in small data scenarios is not our primary focus, it is worth noting that we used a relatively large replay buffer despite the dataset's limited size. The key finding is that the infinite cosine schedule, despite its simplicity, consistently outperforms baselines in both average accuracy and backward transfer (BWT). Notably, the strong performance gains with larger replay buffers suggest that our method scales effectively to large-scale pre-training, where the vast size of modern datasets provides sufficient replay samples to mitigate catastrophic forgetting, even at low buffer sampling rates.

\begin{table*}[t]
    \centering
    \renewcommand{\arraystretch}{1}
    \setlength{\tabcolsep}{5pt}
    \resizebox{0.9\textwidth}{!}{%
    \begin{tabular}{@{}lcc|cc|cc|cc|cc|cc@{}}
        \toprule
        \multirow{3}{*}{\textbf{Task Completed}} 
        & \multicolumn{6}{c|}{\textbf{Acc. $\uparrow$ With ER}} 
        & \multicolumn{6}{c}{\textbf{Acc. $\uparrow$ Without ER}} \\
        \cmidrule(lr){2-7} \cmidrule(lr){8-13}
        & \multicolumn{2}{c|}{ImageNet}& \multicolumn{2}{c|}{Places} & \multicolumn{2}{c|}{FireRisk} 
        & \multicolumn{2}{c|}{{ImageNet}} & \multicolumn{2}{c|}{{Places}} & \multicolumn{2}{c}{{FireRisk}} \\ 
        & {Cos} & {Inf} & {Cos} & {Inf} & {Cos} & {Inf} 
        & {Cos} & {Inf} & {Cos} & {Inf} & {Cos} & {Inf} \\
        \midrule
         ImageNet ($D_0$) & 60.34 & 59.73 & \cellcolor[HTML]{E0E0E0}30.56 & \cellcolor[HTML]{E0E0E0}30.61 & \cellcolor[HTML]{E0E0E0}60.05 & \cellcolor[HTML]{E0E0E0}60.37
                        & 60.34 & 59.73 & \cellcolor[HTML]{E0E0E0}30.56 & \cellcolor[HTML]{E0E0E0}30.61 & \cellcolor[HTML]{E0E0E0}60.05 & \cellcolor[HTML]{E0E0E0}60.37 \\
         Places ($D_1$)    & 58.89 & \textbf{61.09} & 32.35 & 32.03 & \cellcolor[HTML]{E0E0E0}60.28 & \cellcolor[HTML]{E0E0E0}59.68
                        & 49.97 & \textbf{50.77} & 32.26 & 31.95 & \cellcolor[HTML]{E0E0E0}60.13 & \cellcolor[HTML]{E0E0E0}60.58 \\
         FireRisk ($D_2$)  & 54.35 & \textbf{57.50} & 31.12 & \textbf{31.53} & 61.13 & 61.50
                        & 33.39 & \textbf{36.38} & 23.40 & \textbf{25.19} & 62.30 & 62.11 \\
        
        \midrule
        \midrule
        
        \textbf{Metric} & \multicolumn{2}{c|}{\textbf{Avg. Acc.} $\uparrow$} & \multicolumn{2}{c|}{\textbf{FWT} $\uparrow$} & \multicolumn{2}{c|}{\textbf{BWT} $\uparrow$} 
                        & \multicolumn{2}{c|}{\textbf{Avg. Acc} $\uparrow$} & \multicolumn{2}{c|}{\textbf{FWT} $\uparrow$} & \multicolumn{2}{c}{\textbf{BWT} $\uparrow$} \\
        \midrule
        Values & 48.87 & \textbf{50.18} & \textbf{15.51} & 15.23 & -3.61 & \textbf{-1.37} 
               & 39.69 & \textbf{41.22} & 15.43 & \textbf{15.68} & -17.91 & \textbf{-15.06} \\
        \bottomrule
    \end{tabular}
    }
    \caption{\small Performance comparison between cosine (Cos) and infinite cosine (Inf) for MAE pre-training across different tasks, with and without a replay buffer. Grey values indicate performance on datasets that were \textit{unseen} during training at that stage. Each row shows model performance after the model has completed training on the task specified in the row label. The infinite schedule generally preserves knowledge better, particularly in the presence of multiple distribution shifts. Note that this is shown by the superior knowledge retention (bolded) on the previous tasks after learning new tasks.  The bottom section presents key averaged metrics across all three tasks: Average Accuracy (Avg. Acc.), Forward Transfer (FWT), and Backward Transfer (BWT), (where $\uparrow$ indicates that higher is better). Infinite cosine achieves better overall results, especially in reducing forgetting (as shown by less negative BWT values).}
    \label{tab:combined-vision-table}
       \vspace{-0.2cm}
\end{table*}

\subsection{Results for pre-training mae on multiple datasets} 
\label{MAE_fn}
We present the results of our experiments on large scale MAE pre-training in \autoref{tab:combined-vision-table} (left). 
The infinite schedule achieves accuracy comparable to a cosine schedule after ImageNet ($\mathcal{D}_0$) pre-training. 
The effectiveness becomes more pronounced after continual training on Places2 ($\mathcal{D}_1$) with a replay buffer (ER), where the infinite schedule outperforms the cosine schedule on the previous task while achieving better performance on the current dataset. Even under the strong distribution shift introduced by Firerisk ($\mathcal{D}_2$), the infinite cosine schedule proves remarkably robust, achieving \textbf{57.50\%} accuracy on ImageNet. After completing all three tasks, the infinite schedule achieves an average accuracy of \textbf{50.18\%} across all datasets, $\approx$ \textbf{1.3\%} higher than the cosine schedule. The Forward Transfer (FWT) metrics are comparable between the two schedules, while the infinite schedule shows better resistance to catastrophic forgetting with a higher Backward Transfer (BWT). We perform extended forgetting analysis in \autoref{appen:forgetting_metrics}.

Similarly, when evaluating infinite schedule without experience replay in \autoref{tab:combined-vision-table} (right), we observe that it maintains its competitive performance even though there is a significant forgetting. After initial pre-training on ImageNet, it shows comparable performance to the cosine schedule. After pre-training on Places2, infinite schedule demonstrates higher accuracy on the previous task i.e ImageNet. Similar to replay experiment, this is more visible after the third distribution shift where the infinite schedule maintains ImageNet accuracy at \textbf{36.38\%}, outperforming the cosine schedule's \textbf{33.39\%}. This improvement is particularly significant given the challenging nature of continual learning without a replay buffer. In the overall metrics, the infinite schedule achieves a higher average accuracy and Forward Transfer (FWT). Importantly, even without replay, infinite schedule demonstrates better resistance to catastrophic forgetting, with a high Backward Transfer (BWT). 

\begin{figure*}[h!]
    \centering
    \subfloat[Linear probe loss on Imagenet]{\includegraphics[trim={0 0 0 0},clip,width=0.42\textwidth]{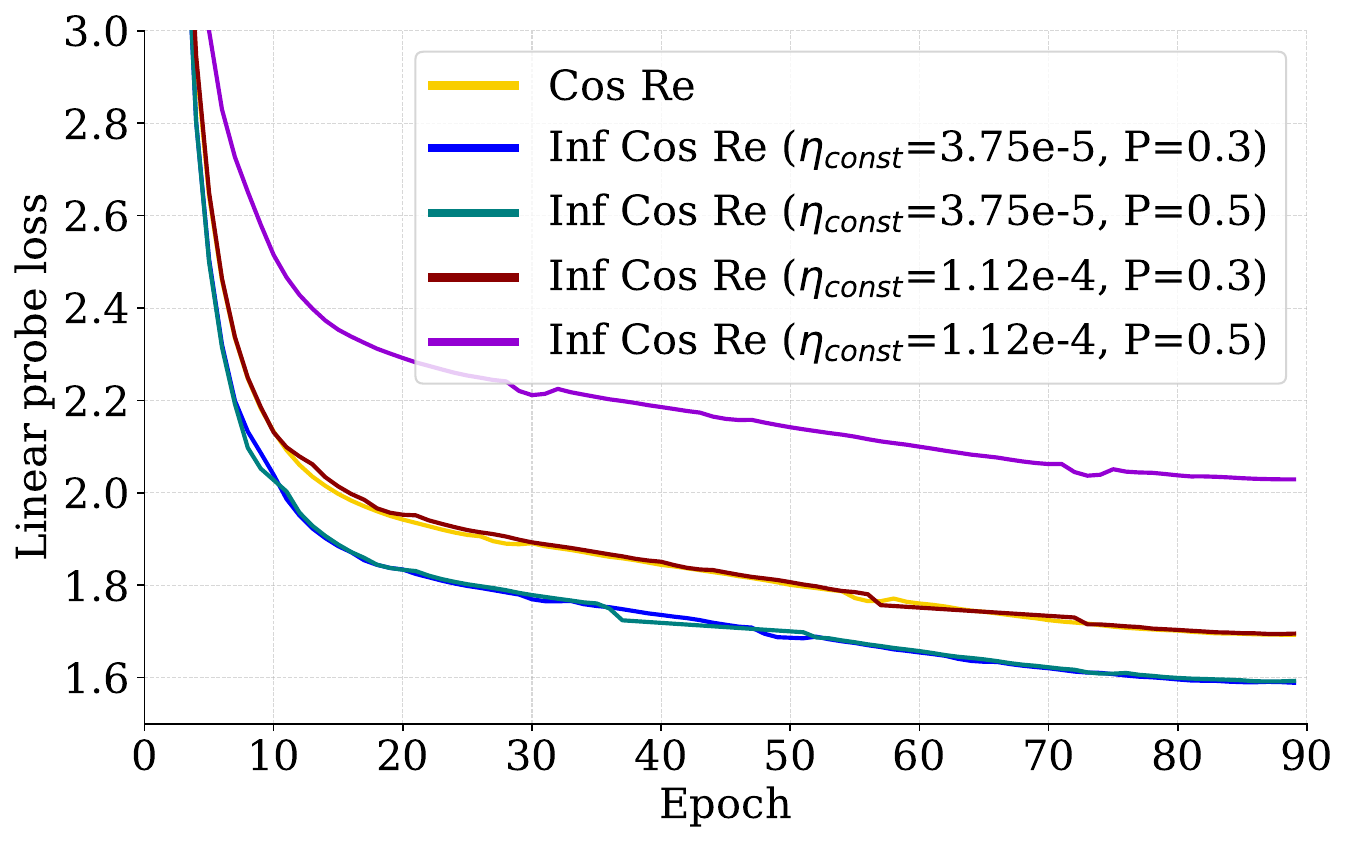}} 
    \subfloat[Linear probe loss on Places2]
    {\includegraphics[trim={27 0 0 0},clip,width=0.4\textwidth]{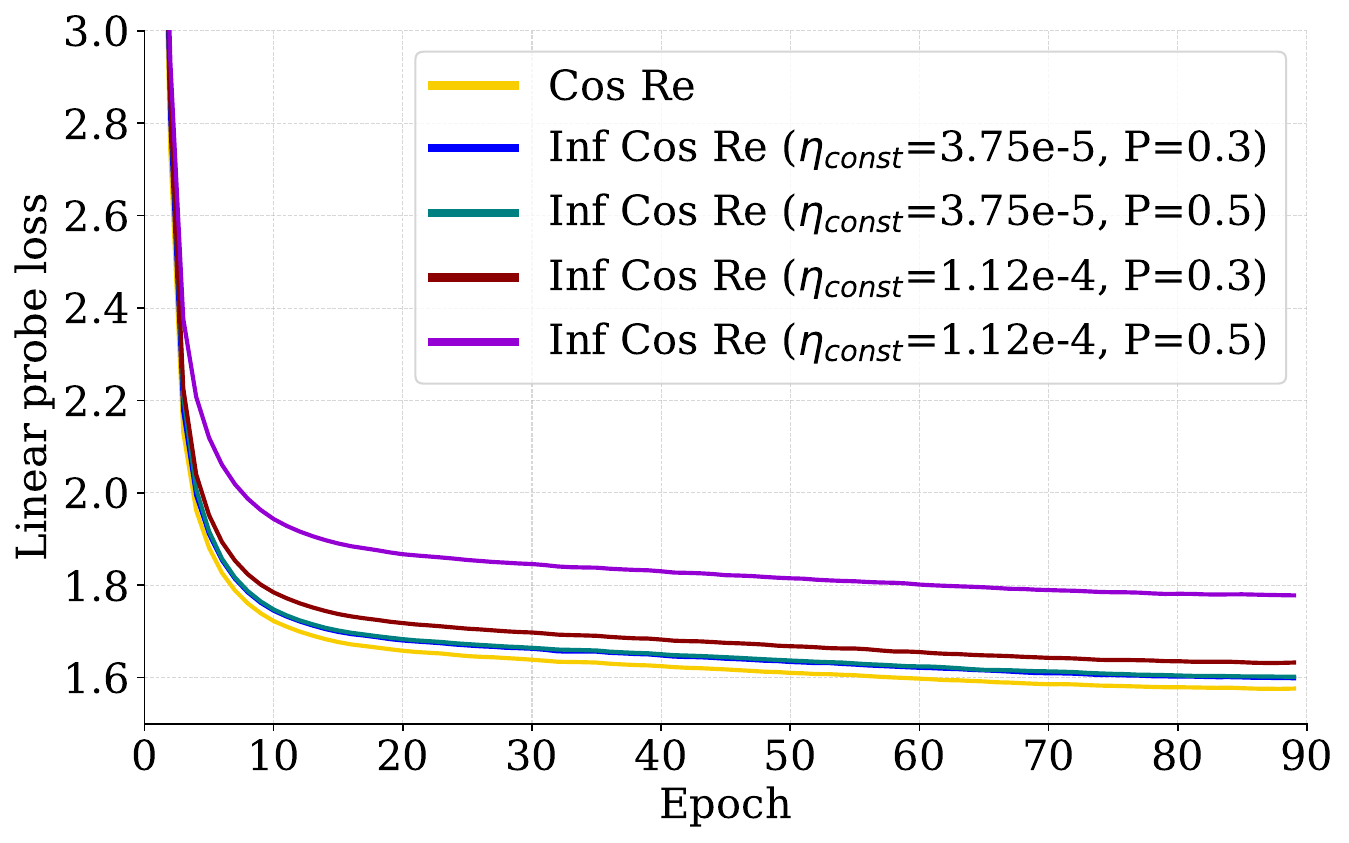}}
    \caption{\small Linear probe loss ($\downarrow$ is better) for cosine schedule and infinite schedule with different constant learning rate and cooldown proportion with replay buffer. We observe that the infinite schedule with a lower $\eta_{const}=3.75e-5$ has the lowest forgetting compared to other schedules.}
    \label{fig:vision_LP_loss_replay}
\end{figure*}

\textbf{Effect of cooldown proportion ($P$) and constant learning rate ($\eta_{const}$):}
In \autoref{fig:vision_LP_loss_replay}, we analyze how the cooldown proportion and constant learning rate in the infinite schedule affect model performance on past and current tasks in ImageNet and Places2. The graphs compare linear probe validation loss across epochs for the standard cosine schedule and infinite schedules with varying configurations. Our analysis shows that lower constant learning rate ($\eta_{const}=3.75e-5$) consistently reduces forgetting as compared to higher rate ($\eta_{const}=1.12e-4$), 
Further, it can be observed that for $\eta_{const}=3.75e-5$ cooldown proportion has negligble effect, but for $\eta_{const}=1.12e-4$ shorter cooldown period ($P=0.3$) outperform longer period ($P=0.5$). This is likely because shorter cooldown phase represents quick decay to a stable $\eta_{const}$ whereas a longer cooldown would mean a high learning rate for longer durations, which could cause instability in training, thus increasing forgetting. 
While the cosine schedule performs better on current tasks, this gap narrows with an appropriately low constant learning rate in the infinite schedule. A similar trend is observed in training without a replay buffer, as shown in \autoref{appn:cooldown-prop}.

\subsection{Results for Continual Pre-training LLMs}
\label{llm_results}
\begin{figure*}[t!]
    \centering
    \vspace{-5pt}
    \subfloat[Valid. Loss on DCLM]{\includegraphics[trim={0 0 0 0},clip, width=0.43 \textwidth]{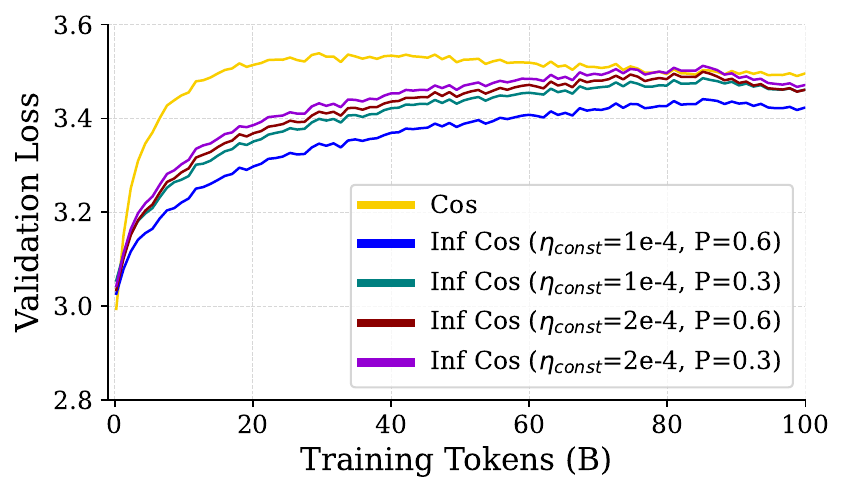}} 
    \subfloat[Valid. Loss on Stack]{\includegraphics[trim={22 0 0 0},clip, width=0.40\textwidth]{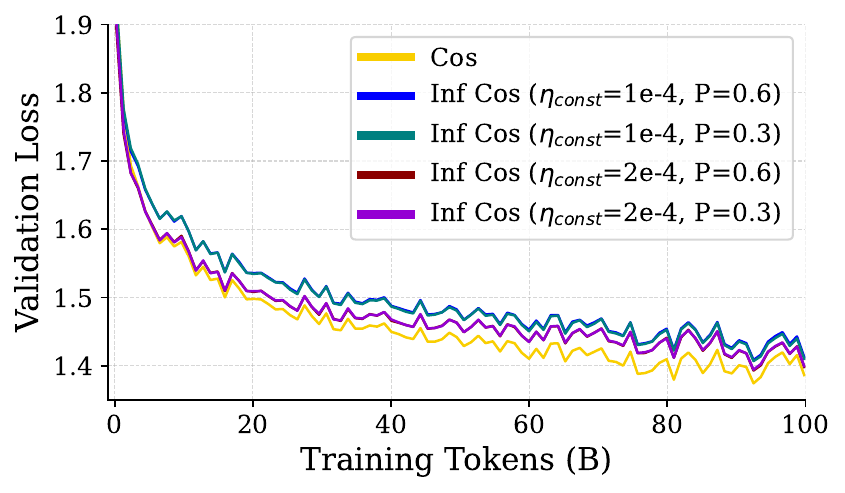}}
    \caption{\small Validation Loss ($\downarrow$ is better) for different schedules. CPT  is on Stack data ($\mathcal{D}_1$), validating on both DCLM ($\mathcal{D}_0$) and Stack ($\mathcal{D}_1$) datasets. All the configurations of infinite schedules mitigate catastrophic forgetting with a lower validation loss on DCLM data, as compared to cosine. However, the downstream performance of infinite schedule on the current task(Stack) is slightly lower than cosine.}
    \label{fig:code_val_loss_without_replay}
       \vspace{-0.3cm}
    \end{figure*} 

\begin{figure*}[t!]
        \centering
        \subfloat[Valid. Loss on DCLM ($50\%$ Replay)]{\includegraphics[trim={0 0 0 0},clip,width=0.43 \textwidth]{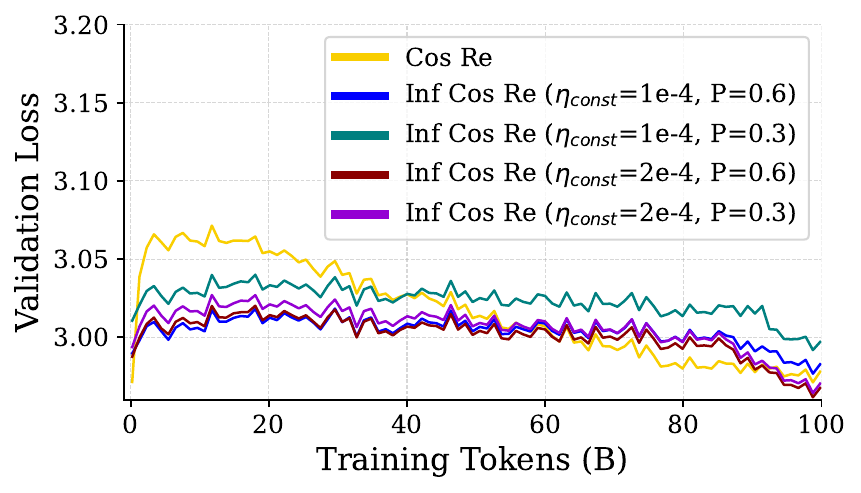}} 
        \subfloat[Valid. Loss on Stack ($50\%$ Replay)]{\includegraphics[trim={22 0 0 0},clip,width=0.40\textwidth]{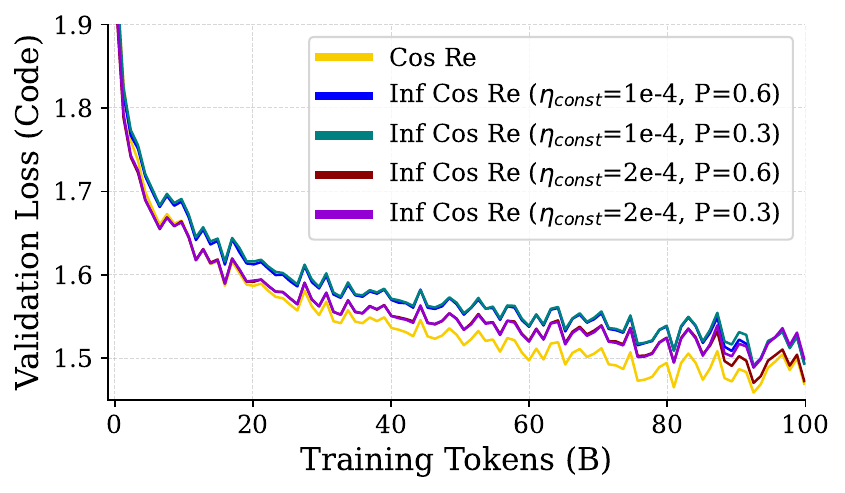}}
        \caption{\small Validation Loss ($\downarrow$ is better) for different schedules accompanied with replay. CPT  is on Stack data ($\mathcal{D}_1$), validating on both DCLM ($\mathcal{D}_0$) and Stack ($\mathcal{D}_1$) datasets. Infinite schedule with $\eta_{const}=2e-4$ and longer cooldown of $P=0.6$ helps in mitigating catastrophic forgetting with minimum validation loss, as compared to cosine and other configurations of infinite scheduling. Even on the current task (Stack), $\eta_{const}=2e-4$ and $P=0.6$ yield a validation loss closely matching that of the cosine schedule.}
        \label{fig:code_val_loss_with_replay}
          \vspace{-0.5cm}
\end{figure*} 
    
We begin pre-training on the DCLM dataset and observe that even in this pre-training phase, rapid annealing in the case of infinite schedule yields a lower validation loss compared to the cosine schedule, offering a competitive advantage. This trend is evident in \autoref{fig:dclm_val_loss}, with further details provided in \autoref{appn:pretrain-dclm}. After pre-training on DCLM, we continue training on the Stack dataset. \autoref{fig:code_val_loss_without_replay} shows the validation loss on the DCLM ($\mathcal{D}_0$) and Stack ($\mathcal{D}_1$) dataset for cosine and infinite schedule with varying $\eta_{const}$ and $P$. We observe that all the configurations of infinite schedule helps in mitigating catastrophic forgetting with a lower validation loss on DCLM data, as compared to cosine, with a minimum validation loss for $\eta_{const}=1e-4$ and longer cooldown of $P=0.6$. This is in concurrence with the observations for MAE large scale pre-training. 

However, we observe that the infinite schedule exhibits slightly lower adaptability to the current task (Stack) compared to the cosine schedule. Specifically, the infinite schedule ($\eta_{const}=1e-4$, $P=0.6$), which minimizes forgetting, shows a marginally higher validation loss on Stack. However, with a higher $\eta_{const}=2e-4$, the infinite schedule achieves performance comparable to cosine on the current task while maintaining a lower validation loss on the upstream task.

To alleviate forgetting, we further introduce a replay mechanism where we sample $50\%$ of the data from the previous task (DCLM) and $50\%$ from the current task (Stack). Figure \autoref{fig:code_val_loss_with_replay} shows the validation loss on the DCLM ($\mathcal{D}_0$) and Stack ($\mathcal{D}_1$) dataset for cosine and infinite schedule with varying $\eta_{const}$ and $P$ with replay. We observe that the infinite schedule with $\eta_{const}=2e-4$ and longer cooldown of $P=0.6$ helps in mitigating catastrophic forgetting with minimum validation loss, as compared to cosine and other configurations of infinite scheduling. We further observe that infinite schedule, irrespective of the $P$ and $\eta_{const}$ gives a lower validation loss as compared to cosine. A higher $\eta_{const}$ likely enhances adaptability to the current task, while a lower $\eta_{const}$ minimizes forgetting on previous tasks. Since replay mitigates forgetting, a higher $\eta_{const}$ ultimately achieves the best overall performance, balancing adaptability and retention. Furthermore, we present experiments in \autoref{appn:dyn_buffer} that demonstrate the flexibility and agility of the infinite cosine schedule in preserving knowledge from previous tasks.

\begin{figure*}[t!]
    \centering
    \subfloat[Valid. Loss on DCLM]{\includegraphics[trim={0 0 0 0},clip, width=0.35\textwidth]{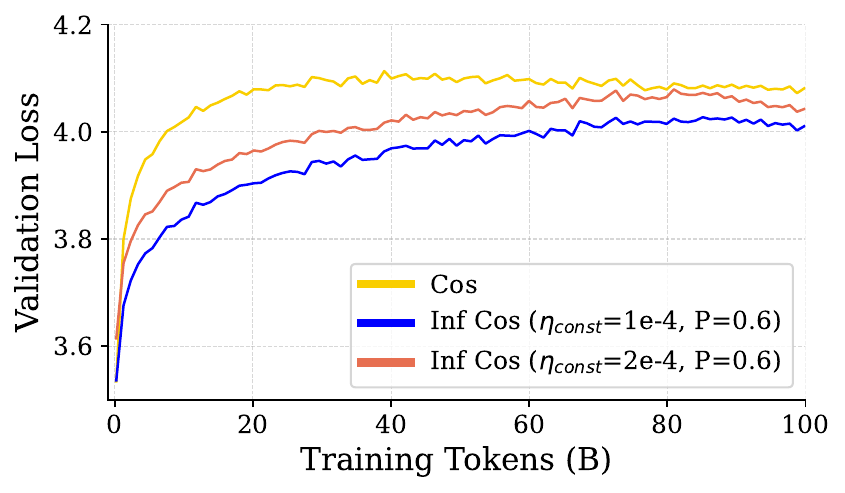}} 
    \subfloat[Valid. Loss on Stack]{\includegraphics[trim={22 0 0 0},clip, width=0.33\textwidth]{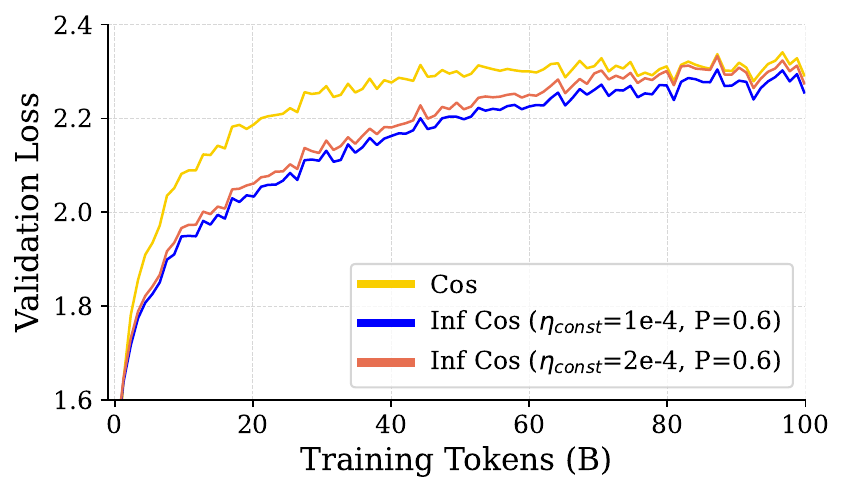}}
    \subfloat[Valid. Loss on German]{\includegraphics[trim={22 0 0 0},clip, width=0.33\textwidth]{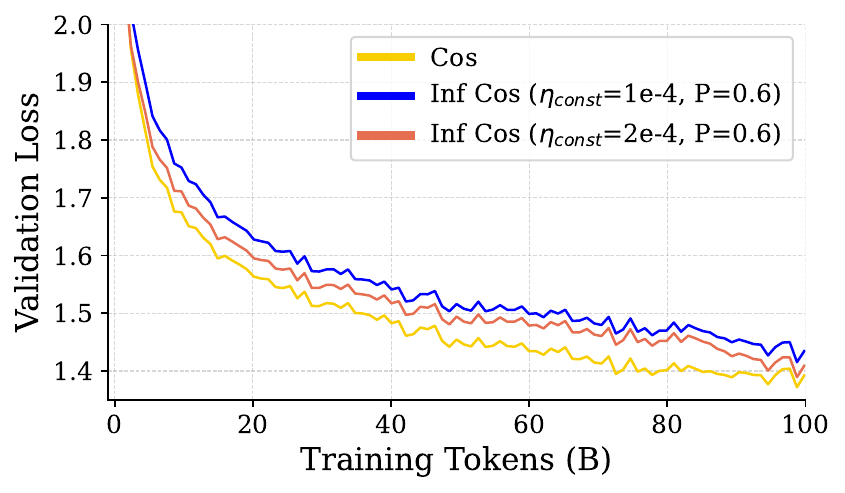}}
    \caption{\small Validation Loss ($\downarrow$ is better) for different schedules. CPT  is on German data ($\mathcal{D}_2$), validating on all German ($\mathcal{D}_2$) DCLM ($\mathcal{D}_0$) and Stack ($\mathcal{D}_1$) datasets. Infinite schedules (both $\eta_{const} \in \{1e-4, 2e-4 \} $) gives a lower validation loss on previous tasks as compared to cosine. The downstream performance of infinite schedule on the current task (German) is comparable to cosine.} \label{fig:german_code_val_loss_without_replay}
    
    \end{figure*}

To further strengthen our evaluation, we introduce a language shift by continually pre-training on the German dataset (German language). This transition imposes a more pronounced distributional shift, as the model moves from programming language data (Stack) to natural language. As in previous sections, we measure validation loss across all datasets while continually pre-training on German as shown in \autoref{fig:german_code_val_loss_without_replay}
Given our earlier findings that short cooldown proportions are detrimental, we train models only with $P=0.6$ under an infinite schedule. Consistent with our previous observations (\autoref{fig:code_val_loss_without_replay}), we find that the infinite schedule with $\eta_{const}=1e-4$ and $P=0.6$ yields the best performance in mitigating forgetting.

While the validation loss provides a good measure of performance on the pre-training objective, LLMs abilities
are typically judged by their performance on evaluation tasks. With the caveat that we use base models, i.e our models have not been instruction-tuned, fine-tuned, or adapted to human preferences in any way, we present their evaluation on popular benchmarks in this section. \autoref{tab:evals_llm} shows the evaluation results on various benchmarks for different schedules. We observe that with replay, the infinite schedule with $\eta_{const}=2e-4$ gives the best performance across all the benchmarks with an average accuracy of \textbf{46.81\%}. For the model after pre-training on German, infinite schedule with $\eta_{const}=1e-4$ gives the best performance across the German evaluation benchmarks with an average accuracy of \textbf{28.10\%} as shown in \autoref{tab:evals_llm_german}. These results highlight that infinite schedules not only circumvent catastrophic forgetting but also provide a competitive advantage in downstream evaluations.

\begin{table*}[t!]
    \centering
    \renewcommand{\arraystretch}{1.2}
    \resizebox{1\textwidth}{!}{
    \begin{tabular}{l l | c c c c c c c |c }
    \toprule
    Scheduler & Training Tokens & LOAI & HS & OBQA & WG & ARC-e & PIQA & LQA & \textbf{Avg.}  \\
    \midrule
    \multirow{3}{*}{Cosine} 
     & 100B DCLM $\cup$ 100B Stack & 46.36 & 42.71 & 32.0 & 50.28 & 51.05 & 68.44 & 27.50 & 45.47 \\
     
    & 100B DCLM $\rightarrow$ 100B Stack & 33.17 & 31.79 & 25.2 & 49.17 & 42.72 & 62.51 & 25.49 & 38.58 \\ 
    
    & 100B DCLM $\rightarrow$ 100B Stack (50\% Replay) & 47.56 & 43.96 & 32.2 & 52.33 & 50.50 & 69.53 & 28.57 & 46.37 \\
    \cmidrule(lr){2-10}
    & 100B DCLM $\rightarrow$ 100B Stack $\rightarrow$ 100B German  & 18.32 & 29.74 & 23.20 & 50.90 & 33.83 & 58.92 & 28.41  & 33.92 \\ 
    
    & 100B DCLM $\rightarrow$ 100B Stack $\rightarrow$ 100B German (50\% Replay) & 44.09 & 40.31 & 30.4 & 53.74 & 48.69 & 67.51 & 27.80 & 44.65 \\
    
    \midrule
    \multirow{4}{*}{Inf Cos ($\eta_{const}=$ 1e-4)} 
    & 100B DCLM $\rightarrow$ 100B Stack & 35.73 & 33.47 & 26.0 & 51.78 & 43.39 & 62.19 & 28.11 & 40.09 \\

    & 100B DCLM $\rightarrow$ 100B Stack (50\% Replay) & 49.16 & 43.72 & 32.60 & 52.09 & 50.59 & 68.93 & 27.65 & 46.39 \\
    \cmidrule(lr){2-10}
    & 100B DCLM $\rightarrow$ 100B Stack $\rightarrow$ 100B German  & 21.42 & 30.27 & 26.80 & 51.22  & 35.60 & 57.56 & 27.96 & 34.66  \\ 
    
    & 100B DCLM $\rightarrow$ 100B Stack $\rightarrow$ 100B German (50\% Replay) & 44.82 & 40.47 & 30.6 & 52.09 & 48.65 & 67.35 & 25.49 & 44.21 \\
    \midrule
    \multirow{4}{*}{Inf Cos ($\eta_{const}=$ 2e-4)} 
    & 100B DCLM $\rightarrow$ 100B Stack & 33.99 & 32.44 & 26.2 & 51.93 & 43.10 & 60.99 & 26.57 & 39.31 \\

    & 100B DCLM $\rightarrow$ 100B Stack (50\% Replay) & 48.73 & 44.42 & 31.6 & 54.85 & 51.73 & 69.31  & 27.04 & \textbf{46.81} \\

    \cmidrule(lr){2-10}
    & 100B DCLM $\rightarrow$ 100B Stack $\rightarrow$ 100B German  & 19.61 & 30.01 & 24.6 & 49.80 & 35.82 & 57.12 & 28.26 & 34.21 \\

     & 100B DCLM $\rightarrow$ 100B Stack $\rightarrow$ 100B German (50\% Replay) & 44.05 & 41.08 & 32.8 & 52.88 & 48.94 & 68.22 & 27.03 & \underline{45.0} \\
     
    \bottomrule
    \end{tabular}}
    \begin{flushleft}\vspace{-2pt}\hspace{8pt}{\fontsize{6}{8}\selectfont LOAI: LambdaOpenAI, HS: HellaSwag, OBQA: OpenBookQA, WG: WinoGrande,LQA: LogicQA}\end{flushleft}
    \caption{\small Zero-shot results on popular LM benchmarks. Normalized accuracy is
    reported. The best performing results (\textbf{Avg.}) are \textbf{bolded} for task DCLM $\rightarrow$ Stack. Similarly, for task DCLM $\rightarrow$ Stack $\rightarrow$ German, best performing results are \underline{underlined}. On average, as expected, we observe that the infinite schedule with $\eta_{const}=2e-4$ with a $50\%$ replay gives the best performance across all the benchmarks. Even without Replay, both the infinite schedules give better performance as compared to cosine. This demonstrates the effectiveness of the infinite schedule in mitigating forgetting.}
    \label{tab:evals_llm}
\end{table*}

\begin{table*}[h!]
    \centering

    \resizebox{0.74\textwidth}{!}{
    \begin{tabular}{l l | c c |c }
    \toprule
    Scheduler & Training Tokens & ARC-de & HS-de & \textbf{Avg.}  \\
    \midrule
    \multirow{1}{*}{Cosine} 
    

    & 100B DCLM $\rightarrow$ 100B Stack $\rightarrow$ 100B German & 23.29 & 32.89  & 28.09 \\ 

     
    \midrule
    \multirow{1}{*} {Inf Cos ($\eta_{const}=$ 1e-4)}

    & 100B DCLM $\rightarrow$ 100B Stack $\rightarrow$ 100B German & \textbf{23.64} & 32.56 & \textbf{28.10}  \\ 
    \midrule
    \multirow{1}{*}{Inf Cos ($\eta_{const}=$ 2e-4)} 
    & 100B DCLM $\rightarrow$ 100B Stack $\rightarrow$ 100B German & 23.21 & \textbf{32.90} & 28.06\\ 

    \bottomrule
    \end{tabular}}
    
    \caption{\small Zero-shot results showing adaptability of the model after completing training on the German data ($\mathcal{D}_2$) on popular LM benchmarks. We observe that the infinite schedule with $\eta_{const}=1e-4$ achieves the best performance on German evaluation benchmarks, demonstrating that the infinite schedule adapts more effectively than the cosine schedule on the most recent task.
}
    \label{tab:evals_llm_german}
\end{table*}

\section{Discussion}
\label{sec:disc}
In our large-scale experiments, we have explored different hyperparameters of the infinite cosine schedule across both vision and language tasks. In the case without replay, the choice of $\eta_{const}$ follows a similar pattern across both modalities, where a lower $\eta_{const}$ yields optimal performance. However, with replay, an apparent discrepancy emerges: vision tasks still favor a lower $\eta_{const}$, while language tasks seem to benefit from a higher $\eta_{const}$. 
In vision tasks, the variation between high and low $\eta_{const}$ spans an order of magnitude (i.e., a factor of 10x), whereas in language tasks, the difference is narrower. This suggests that the relative comparison of $\eta_{const}$ across modalities is not directly meaningful, as the scales of sensitivity differ between vision and language models. 
Consequently, we say that optimal
constant learning rate should be selected through careful hyperparameter tuning which is one limitation of our work.


\section{Conclusion}

Our results suggest that infinite cosine schedules offer a flexible and robust framework for continual pre-training (CPT) of foundation models across vision and language domains. They enable seamless training continuation from intermediate checkpoints (e.g., at $\eta_{const}$), support dynamic adaptation strategies such as adjustable replay, and maintain strong performance under distribution shifts, without requiring a predefined training budget. We saw that infinite schedules allow us to seamlessly resume training (continued pre-training at $\eta_{const}$), that alone or along with replay, they best continual learning baselines, and, on large-scale experiments across multiple vision and language datasets, they consistently outperform repeated cosine decay. While we do not claim universal superiority, our large-scale experiments demonstrate that infinite schedules consistently provide competitive retention of prior knowledge and improved stability in non-IID continual learning scenarios, making them a practical alternative to repeated cosine decay in real-world CPT pipelines.

Our exploration of infinite learning rate schedules opens promising avenues for future research. For example, exploring the theoretical underpinnings of infinite schedules to establish a more rigorous foundation for their effectiveness in continual pre-training, comparing different cooldown functions across modalities and extending these studies to a wider range of architectures and self-supervised learning frameworks are all important directions for future work.

\newpage

\section{acknowledgements}
We acknowledge support from NSERC Discovery Grant RGPIN- 2021-04104 [E.B.], FRQNT New Scholar [\emph{E.B.}], the Canada CIFAR AI Chair Program [I.R.], and the Canada Excellence Research Chairs Program [I.R.]. We would also like to acknowledge funding from the FRQNT Doctoral (B2X) scholarship [B.T.]. This research was made possible thanks to the computing resources on the Frontier supercomputer, provided as a part of the ALCC 2024 program award ``Scalable Foundation Models for Transferable Generalist AI”.  These resources were provided by the Oak Ridge Leadership Computing Facility at the Oak Ridge National Laboratory, which is supported by the Office of Science of the U.S. Department of Energy under Contract No. DE-AC05-00OR22725. In particular, we thank Jens Glaser for his help with the Summit supercomputer. We would also like to thank Isaac and Nafi of ORNL for their assistance in obtaining the MRI dataset used in additional experiments.

\bibliography{ref}
\bibliographystyle{collas2025_conference}

\clearpage
\appendix

\section{Implementation details and hyperparameters for Vision pre-training}
\label{appen:implementation_vision}

\subsection{Formal Definition of MAE Pre-training}
\label{appn:mae-pretrain}

Formally the MAE pre-training procedure is described as follows: For each image $\mathbf{x} \in \mathcal{D}$, where $\mathcal{D} = \{\mathbf{x}_i\}_{i=1}^{N}$ and $\mathbf{x}_i \sim \text{IID}$, we first partition it into a sequence of non-overlapping patches $\{\mathbf{p}_i\}_{i=1}^N$. We use the same masking ratio from the original MAE~\citep{he_masked_2022}  that randomly masks 75\% of these patches, creating two complementary sets: visible patches $\mathcal{V}$ and masked patches $\mathcal{M}$. An encoder $f_\theta(\cdot)$, implemented as a Vision Transformer \citep{dosovitskiy2020image}, processes only the visible patches to obtain latent representations $\mathbf{h}_v = f_\theta(\{\mathbf{p}_i\}_{i \in \mathcal{V}})$. These encoded features, along with  mask tokens $\{\mathbf{m}_j\}_{j \in \mathcal{M}}$, are then fed to a decoder $g_\phi(\cdot)$ to reconstruct the original image: $\hat{\mathbf{x}} = g_\phi(\{\mathbf{h}_v\} \cup \{\mathbf{m}_j\})$. The entire framework is trained end-to-end by minimizing the mean squared error loss $\mathcal{L}_{mse} = \|\mathbf{x} - \hat{\mathbf{x}}\|_2^2$ between the original and reconstructed images. After pre-training, the decoder is discarded, and the encoder serves as a feature extractor for downstream vision tasks.

\subsection{Hyperparameters and Implementation details for CIFAR10 MAE}
\label{appn:hyp-vision-tiny}

For our architecture, we employ a ViT-tiny encoder (12 layers, 192 hidden dimensions, 3 attention heads) to match the scale of CIFAR-10, with our implementation based on \citet{mae_cifar}'s work. Our model uses a masking ratio of 0.75, consistent with the original MAE, and incorporates a lightweight decoder (4 layers) with learned position embeddings to reconstruct the masked patches. Regarding the learning rate configuration, we selected a maximum learning rate of 7.5e-5 through hyperparameter tuning over the values [7e-5, 1.5e-4, 3e-4] on the first two tasks, with a minimum learning rate of 7.5e-6. For most experiments, we employ a constant learning rate of 1.875e-5 and a cooldown proportion of 0.4, except for experiments without replay, where we increase the constant learning rate to 5.625e-5. These optimal values were determined through experiments similar to our large-scale setup, testing cooldown proportions [0.3, 0.4, 0.5] and constant learning rates [1.875e-5, 5.625e-5]. While these findings align with our large-scale experiments, the small dataset size necessitated a slightly larger cooldown proportion to maintain a higher learning rate for a longer duration. For linear probing experiments in our small-scale setup, we utilized the AdamW \citep{adamw} optimizer with a weight decay coefficient of $5e-3$ and momentum parameters $\beta_1$ and $\beta_2$ set to 0.9 and 0.95, respectively. The linear probing experiments implemented a cosine decay learning rate schedule with a maximum learning rate $\eta_{max}=1e-3$, running for 100 epochs total, including 10 warmup epochs, with a batch size of 128. Complete hyperparameter details for pre-training and linear probing can be found in the corresponding tables \autoref{tab:hyp-pretrain-small-vision} and \autoref{tab:hyp-linprobe-vision-small}. 
For the baseline methods MAS \citep{aljundi2018memory} and LwF \citep{Li17learning}, we conducted hyperparameter tuning using grid search over the first two tasks. For MAS, we explored values of $\alpha$ and $\lambda$ in [0.25, 0.5, 0.75]. Similarly for LwF, we searched for optimal $\alpha$ values within the same range.
\begin{table}[ht]
\begin{minipage}{0.48\textwidth}
\centering
\begin{tabular}{@{}cc@{}}
\toprule
Description    & Value    \\ \midrule
optimizer      & AdamW    \\
weight decay   & 5.00e-03 \\
$\beta_1$      & 0.9      \\
$\beta_2$      & 0.95     \\
batch size     & 512      \\
warmup epochs  & 20       \\
Total epochs   & 400      \\
Max learning rate $\eta_{max}$   & 7.50e-05 \\
Min learning rate $\eta_{min}$   & 1.50e-06 \\
Constant learning rate $\eta_{const}$ & 1.875e-5 \\ 
\midrule 
\multicolumn{2}{c}{\textbf{ViT-tiny}}\\
Parameters & $7M$ \\
Num Attention Heads & $3$ \\
Num Layers & $12$ \\
Hidden Size & $192$ \\
Hidden Activation & GeLU \\
Positional Embedding & Learnable \\
Patch Size & $2 \times 2$ \\
Image Size & $32 \times 32$ \\
Dropout Rate & $0.1$ \\
\bottomrule
\end{tabular}
\caption{Hyperparameters for pre-training on the small scale setup}
\label{tab:hyp-pretrain-small-vision}
\end{minipage}%
\hfill
\begin{minipage}{0.48\textwidth}
\centering
\begin{tabular}{@{}cc@{}}
\toprule
Description            & Value        \\ \midrule
optimizer              & AdamW        \\
weight decay           & 5.00e-03     \\
$\beta_1$              & 0.9          \\
$\beta_2$              & 0.95         \\
learning rate schedule & cosine decay \\
batch size             & 128          \\
warmup epochs          & 10           \\
Total epochs           & 100           \\
$\eta_{max}$           & 1.00e-03     \\ \bottomrule
\end{tabular}
\caption{Hyperparameters for linear probing on the small scale setup.}
\label{tab:hyp-linprobe-vision-small}
\end{minipage}
\end{table}

\begin{figure*}[ht!]
    \centering
    \subfloat[Linear probe loss on Imagenet]{\includegraphics[trim={0 0 0 0},clip,width=0.48\textwidth]{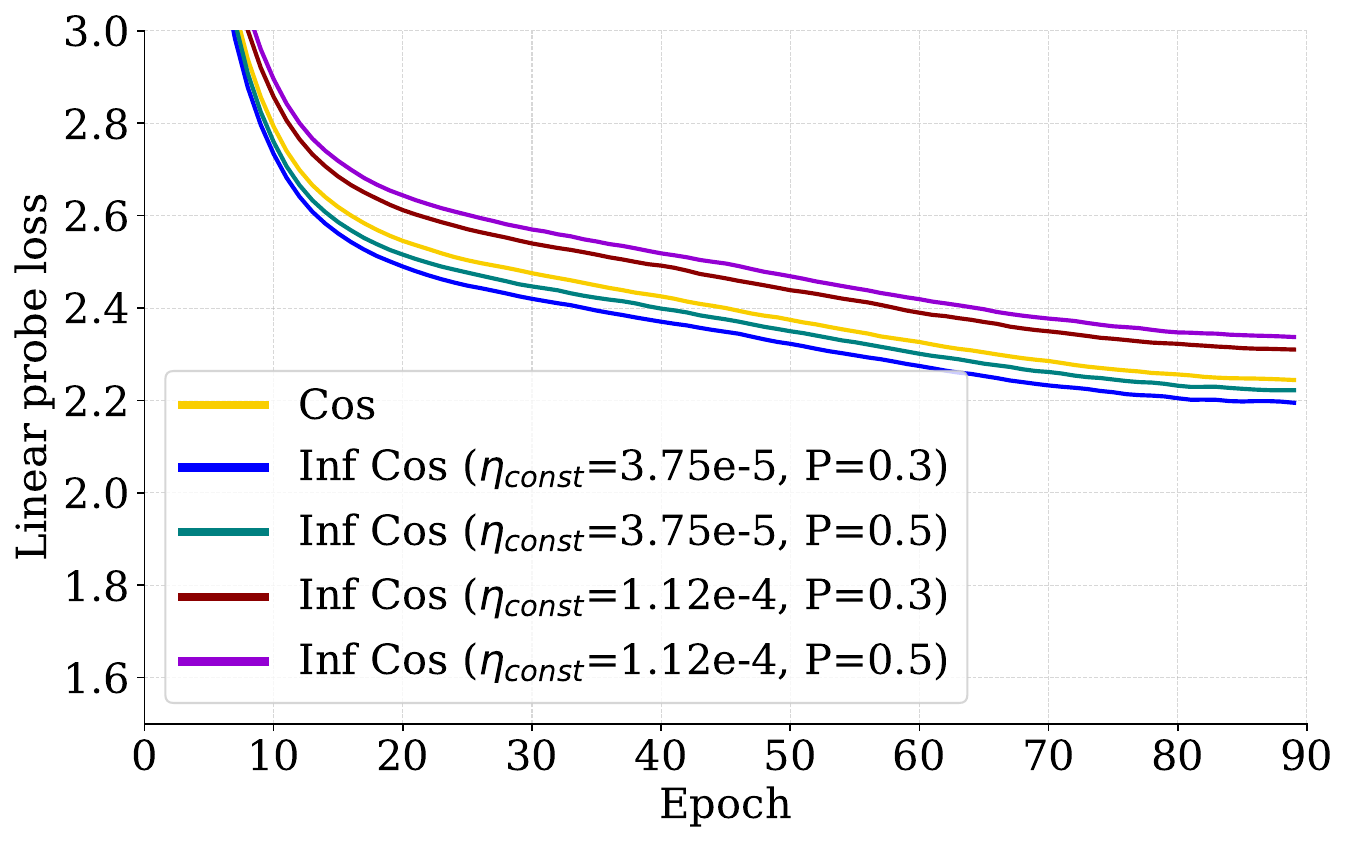}} 
    \subfloat[Linear probe loss on Places2]{\includegraphics[trim={0 0 0 0},clip,width=0.48\textwidth]{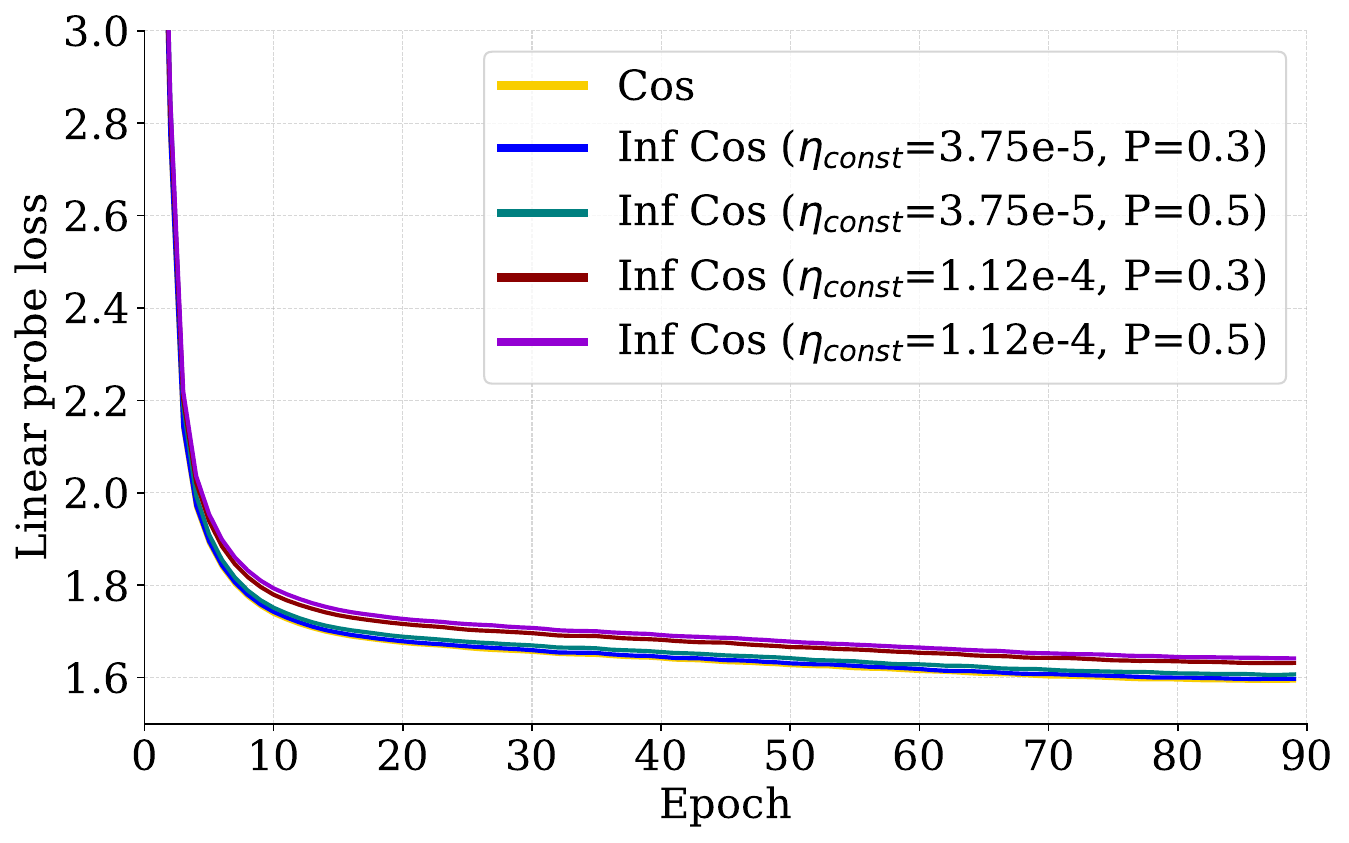}}
    \caption{Linear probe loss ($\downarrow$ is better) for cosine scheduler and infinite scheduler with different configurations without replay buffer. Infinite learning schedule with lower constant learning rate has lower forgetting compared to cosine schedule}
    \label{fig:vision_LP_loss_without_replay}
\end{figure*}

\subsection{Evaluation Metrics for MAE CPT}
\label{appn:eval-metrics}
For evaluation, we employ three key metrics following~\citep{lopez2017gradient}.\textbf{ Average Accuracy} ($Acc = \frac{1}{T}\sum_{i=1}^{T} R_{T,i}$) provides an overall measure of model performance across all tasks, where $T$ is the total number of tasks and $R_{T,i}$ represents the performance on task $i$ after training on all $T$ tasks.\textbf{ Forward Transfer} ($FWT = \frac{1}{T-1}\sum_{i=2}^{T} (R_{i-1,i} - b_{i})$) measures the model's ability to leverage knowledge from previous tasks, where $b_i$ represents the accuracy of a randomly initialized feature extractor. \textbf{Backward Transfer} ($BWT = \frac{1}{T-1}\sum_{i=1}^{T-1} (R_{T,i} - R_{i,i})$) quantifies the impact of subsequent task learning on previous task performance.

\subsection{Implementation details MAE on Imagenet, Places , Firerisk}
\label{appn:mae-pretrain-large-scale}
Our implementation builds upon the PyTorch \citep{paszke2019pytorch} implementation of MAE \citep{he_masked_2022} with ViT-B/16~\citep{he_masked_2022} backbone architecture with 12 layers, 768 hidden dimension, and 12 attention heads. For the infinite learning rate schedule, we maintain a constant learning rate $\eta_{const}=$3.75e-5 during constant phase, while our baseline employs the standard cosine decay schedule.To ensure fair comparison, both schedules share identical maximum $\eta_{max}=1.5e-04$ and minimum  learning rate, $\eta_{min}=1.5e-06$, with hyperparameters for the cosine schedule directly adopted from \citet{he_masked_2022}. We also employ learning rate scaling similar to \citet{goyal2018accurate}. We list all the hyperparameters on \autoref{tab:hyp-pre-large-vision}. To mitigate catastrophic forgetting, we implement a replay buffer with a buffer size of $B = 0.05 \times |\mathcal{D}_i|$ per task, utilizing uniform random sampling for buffer updates. All experiments utilize the AdamW optimizer, \citep{adamw} with training conducted over 300 epochs per task. Following, \citet{ibrahim2024simple} we reset the optimizer states before each task. For linear probing as shown in \autoref{tab:hyp-linprob-large-vision}, we utilized the LARS optimizer with no weight decay ($\lambda=0$). The optimizer's momentum parameter $\beta_1$ was set to 0.9. The learning rate followed a cosine decay schedule with a maximum learning rate ($\eta_{max}$) of $1.00\times10^{-1}$. Training was conducted over 90 epochs with a large batch size of 4096 and included RandomResizedCrop augmentation. This configuration leverages the LARS optimizer's efficiency for large-batch training while maintaining training stability across the diverse image datasets.
\begin{table}[ht]
\begin{minipage}{0.48\textwidth}
\centering
\begin{tabular}{@{}cc@{}}
\toprule
Description    & Value             \\ \midrule
optimizer      & AdamW             \\
weight decay   & 0.05              \\
$\beta_1$      & 0.9               \\
$\beta_2$      & 0.95              \\
batch size     & 4096              \\
warmup epochs  & 40                \\
augmentation   & RandomResizedCrop \\
Total epochs   & 300               \\
Max learning rate $\eta_{max}$   & 1.50e-04          \\
Min learning rate $\eta_{min}$   & 1.50e-06          \\
Constant learning rate $\eta_{const}$ & 3.75e-05          \\ 
\midrule 
\multicolumn{2}{c}{\textbf{ViT-B/16}}\\
Parameters & $86M$ \\
Num Attention Heads & $12$ \\
Num Layers & $12$ \\
Hidden Size & $768$ \\
Hidden Activation & GeLU \\
Weight Decay & $0.3$ \\
Positional Embedding & Learnable \\
Patch Size & $16 \times 16$ \\
Image Size & $224 \times 224$ \\
Dropout Rate & $0.1$ \\

\bottomrule
\end{tabular}
\caption{Hyperparameters for pre-training MAE on Imagenet, Places and Firerisk}
\label{tab:hyp-pre-large-vision}
\end{minipage}%
\hfill
\begin{minipage}{0.48\textwidth}
\centering
\begin{tabular}{@{}cc@{}}
\toprule
Description            & Value             \\ \midrule
optimizer              & LARS              \\
weight decay           & 0                 \\
$\beta_1$              & 0.9               \\
learning rate schedule & cosine decay      \\
batch size             & 4096              \\
warmup epochs          & 10                \\
augmentation           & RandomResizedCrop \\
Total epochs           & 90                \\
$\eta_{max}$           & 1.00e-01          \\ \bottomrule
\end{tabular}
\caption{Hyperparameters for linear probing on ImageNet, Places and Firerisk}
\label{tab:hyp-linprob-large-vision}
\end{minipage}
\end{table}

\section{Effect of cooldown proportion and constant learning rate}
\label{appn:cooldown-prop}

Our analysis in \autoref{fig:vision_LP_loss_without_replay} (a) and (b) investigates learning dynamics in scenarios without a replay buffer, comparing the standard cosine schedule against infinite schedules through linear probe validation loss across epochs. The results mirror patterns observed with replay mechanisms, albeit with substantially higher catastrophic forgetting. Lower constant learning rates ($\eta_{const}$=3.75e-5) exhibit markedly reduced forgetting compared to higher rates ($\eta_{const}$=1.12e-4).
For the lower constant learning rate, we observe that cooldown proportion has minimal impact on performance. In contrast, with higher constant learning rates, shorter cooldown periods yield better performance than longer ones. The dramatic increase in forgetting without replay underscores the critical importance of replay mechanisms in preserving cross-task performance.

\section{Implementation details and hyperparameters for language pre-training}
\label{appn:impl-language}
All models are trained with AdamW \citep{adamw} on 100B tokens for each dataset, using a batch size of 1024 and a sequence length of 2048 approximately corresponding to $47,684$ total training steps. Optimizer states get reset between datasets, as this is common when we have to begin from an open weight model (e.g. from Huggingface \citep{huggingface-transformers}). 
We train with data parallelism across 32 nodes, each equipped with 8 GPUs, maintaining a micro-batch size of 4. The training setup includes activation checkpointing \citep{chen2016training} and ZeRO-1 optimizer sharding \citep{rajbhandari2020zero} to reduce memory overhead. 
\begin{table*}[h!]
    \centering
    \begin{minipage}[t]{0.48\textwidth}
    \caption{\textbf{Hyperparameters of LR schedules.} All models used the same LR schedule hyperparameters. We refer the readers to~\citep{ibrahim2024simple} section 7.2 for a more thorough explanation of these schedules.}
    \label{table:hyperparameters-schedules}
    \centering
    \begin{tabular}{ll}
    \toprule
    Description & Value\\\midrule
    \multicolumn{2}{l}{\textbf{Pre-training}}\\
    Total Iterations & 47684 \\
    Max learning rate ($\eta_\textit{max}$) & $3\cdot10^{-4}$ \\
    Min learning rate ($\eta_\textit{min}$) & $3\cdot10^{-5}$ \\
    Constant learning rate ($\eta_\textit{const}$) & $1\cdot10^{-4}$ \\
    Warmup percent ($N_\textit{w}$) & $1$ \\
    Cooldown iters percent ($N_\textit{c}$) & $60$ \\
    Constant iters percent ($N_\textit{d}$) & $25$ \\\midrule
    \multicolumn{2}{l}{\textbf{Continual Pre-training}}\\
    Total Iterations & $47684$\\
    Max learning rate ($\eta_\textit{max}$) & $3\cdot10^{-4}$ \\
    Min learning rate ($\eta_\textit{min}$) & $3\cdot10^{-5}$ \\
    Constant learning rate ($\eta_\textit{const}$) & $1\cdot10^{-4}$ \\
    Warmup percent ($N_\textit{w}$) & $1$ \\
    Cooldown iters percent ($N_\textit{c}$) & $0$ \\
    Constant iters percent ($N_\textit{d}$) & $85$ \\
    \bottomrule
    \end{tabular}
    \end{minipage}
    \hfill
    \begin{minipage}[t]{0.48\textwidth}
    \centering
    \vspace{2pt}
    \caption{\textbf{Hyperparameters of the ViT and LM transformers in our study.} }
    \label{table:hyperparameters}
    \begin{tabular}{ll}
    \toprule
    Description & Value\\\midrule
    \multicolumn{2}{l}{\textbf{Dense Transformer LM}}\\
    Parameters&$571,148,288$ \\ 
    Non-Embedding Parameters&$439,814,144$ \\
    Num attention heads & 16 \\\midrule
    Num layers& 24\\
    Hidden size & 1024\\
    FFN Hidden size & 2816\\
    FFN Type & GeGLU\\
    Optimizer & AdamW\\
    $\beta_1$,$\beta_2$ &$0.9,0.95$\\
    Batch size & $1024$\\
    Sequence length & $2048$\\
    Hidden activation & GeLU\\
    Weight decay & $0.1$ \\
    Gradient clipping & $1.0$\\
    Decay & Cosine\\
    Positional embedding & Rotary \\
    GPT-J-Residual & True \\
    Weight tying & False\\
    Vocab Size & 128000 \\
    Rotary PCT &0.25 \\
    \multicolumn{2}{l}{\textbf{ViT-B/16}}\\
Parameters & $86,567,656$ \\
Num Attention Heads & $12$ \\\midrule
Num Layers & $12$ \\
Hidden Size & $768$ \\
FFN Hidden Size & $3072$ \\
FFN Type & MLP \\
Optimizer & Adam \\
$\beta_1, \beta_2$ & $0.9, 0.999$ \\
Batch Size & $4096$ \\
Sequence Length & $197$ \\
Hidden Activation & GeLU \\
Weight Decay & $0.3$ \\
Gradient Clipping & $1.0$ \\
Positional Embedding & Learnable \\
Patch Size & $16 \times 16$ \\
Image Size & $224 \times 224$ \\
Dropout Rate & $0.1$ \\
    \multicolumn{2}{l}{\textbf{Common}}\\
    \bottomrule
    \end{tabular}
    \end{minipage}
\end{table*}

\section{Pretraining with DCLM data}
\label{appn:pretrain-dclm}
\begin{figure*}[h!]
    \centering
    {\includegraphics[trim={0 0 0 0},clip,width=0.5\textwidth]
    {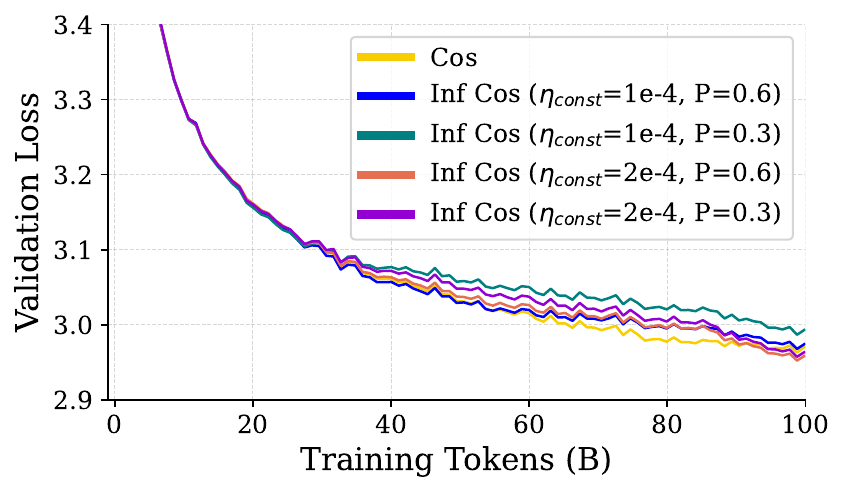}}
    \caption{Validation Loss($\downarrow$ is better) for Different schedules, Training and Validating on DCLM Dataset}
    \label{fig:dclm_val_loss} 
    \end{figure*} 

Figure \ref{fig:dclm_val_loss} shows the validation loss on the DCLM dataset for cosine and infinite schedule with varying $\eta_{const}$ and cooldown proportion $P$. We observe that the infinite schedule with a higher constant learning rate ($\eta_{const}=2e-4$) and cooldown proportion ($P=0.6$) performs better than the cosine schedule and the other configurations of the infinite schedule. The final checkpoint, in the case of infinite schedule, is obtained via annealing which we perform for $15\%$ of the total iterations after the constant phase, as shown in \autoref{fig:cosine_inf_schedule}. It can be inferred that the infinite schedule with $\eta_{const}=2e-5$ and $P=0.6$ performs the best, with validation loss rapidly decaying in the annealing phase. We also observe that a shorter cooldown phase ($P=0.3$) results in suboptimal performance with higher validation loss, thus indicating that a longer cooldown phase is beneficial. We note that this corresponds to $28K$ steps.
As for the $\eta_{const}$, we observe that both $1e-4$ and $2e-4$ perform similarly, with the latter having a slightly lower validation loss, indicating that a higher constant learning rate gives a better exploration possibility during training. 

\begin{figure*}[t!]
    \centering
    {\includegraphics[trim={0 0 0 0},clip,scale=0.6]
    {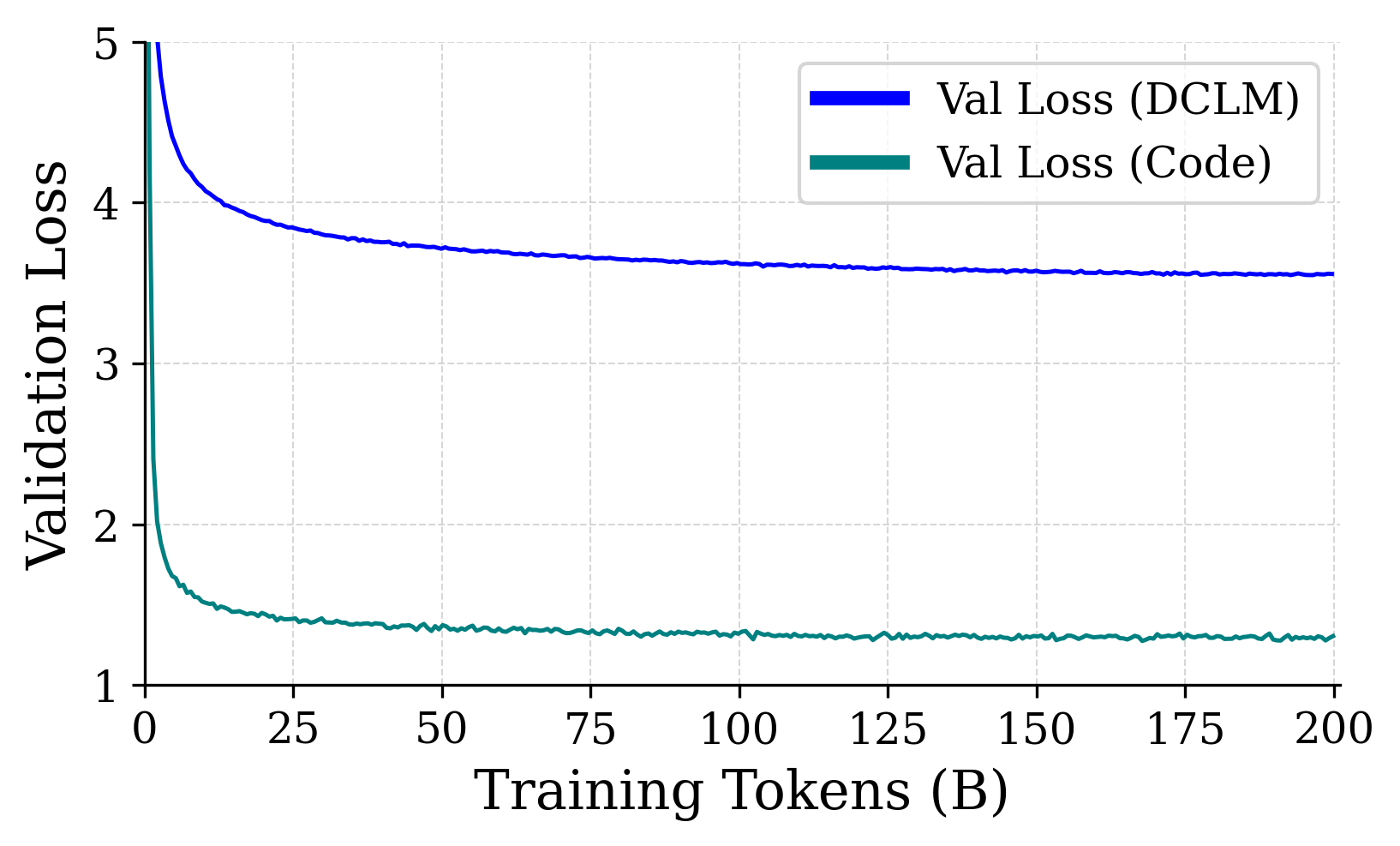}}
    \caption{Validation Loss($\downarrow$ is better) for Cosine while training on combined DCLM and Code}
    \label{fig:dclm_val_loss_combined}
\end{figure*}

\section{Pre-training with combined DCLM and Stack Data}
\label{combined_dclm_stack}
We show the validation loss on the combined DCLM and Stack dataset with cosine scheduling in Figure \ref{fig:dclm_val_loss_combined}. It can be inferred that both the validation loss on DCLM and Stack is worse as compared to continual pre-training with infinite learning schedule. This indicates that the infinite schedule is able to preserve the knowledge of the previous task as well as improve transferability better as compared to cosine schedule, even with combined training on both tasks. 
\begin{figure*}[h!]
    \centering
    \subfloat[Valid. Loss on DCLM]{\includegraphics[trim={0 0 0 0},clip, width=0.35\textwidth]{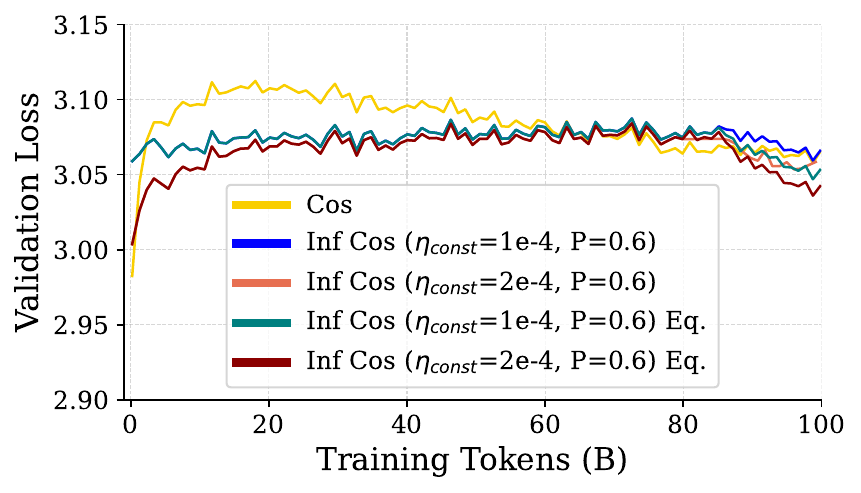}} 
    \subfloat[Valid. Loss on Stack]{\includegraphics[trim={22 0 0 0},clip, width=0.33\textwidth]{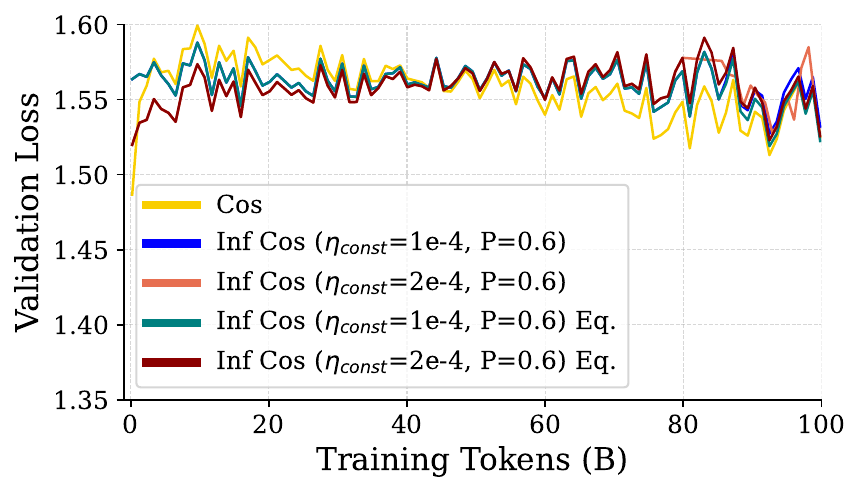}}
    \subfloat[Valid. Loss on German]{\includegraphics[trim={22 0 0 0},clip, width=0.33\textwidth]{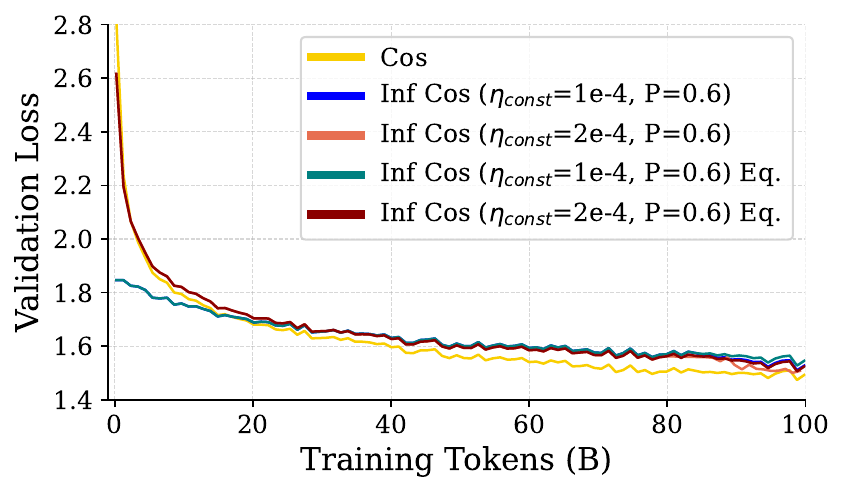}}
    \caption{\small Validation Loss ($\downarrow$ is better) for different schedules accompanied with replay. The total fraction of replay is 50\%, with 25\% of DCLM and other 25\% from Stack Data. CPT  is on German data ($\mathcal{D}_2$), validating on all German ($\mathcal{D}_2$) DCLM ($\mathcal{D}_0$) and Stack ($\mathcal{D}_1$) datasets. We further perform an experiment with equal proportion of data during annealing, referred to as Eq. in the above graphs. It can be observed that equal proportions improve the upstream performance (lower validation loss on DCLM and Stack). The downstream performance is quite similar to other Infinite schedules (both $\eta_{const} \in \{1e-4, 2e-4 \} $) config. The downstream performance of the infinite schedule on the current task (German) is comparable to cosine.} \label{fig:german_code_val_loss_with_replay_eq_ann}
    
    \end{figure*}


\section{Dynamic Adjustment of Replay Buffer with equal proportion of pre-training data during Annealing Phase}
\label{appn:dyn_buffer}
We perform an ablation by annealing on equal proportion of data, i.e. 33.33\% each of DCLM ($\mathcal{D}_0$), Stack ($\mathcal{D}_1$) and German ($\mathcal{D}_2$). As shown in \autoref{fig:german_code_val_loss_with_replay_eq_ann}, it can be inferred that the performance on previous tasks improve as compared to the config where we use 50\% of buffer, since now the proportion of previous tasks data has increased (25\% for each of DCLM and Stack $\rightarrow$ 33.33\%). But this does not deteriorate the downstream performance on German data. Hence in cases, where upstream performance is critical, we can anneal on an equal proportion of data.

\section{Effect of Checkpoint selection in Past task performance}

\begin{figure}[h!]
    \centering
    
    {\includegraphics[width=0.42\textwidth]{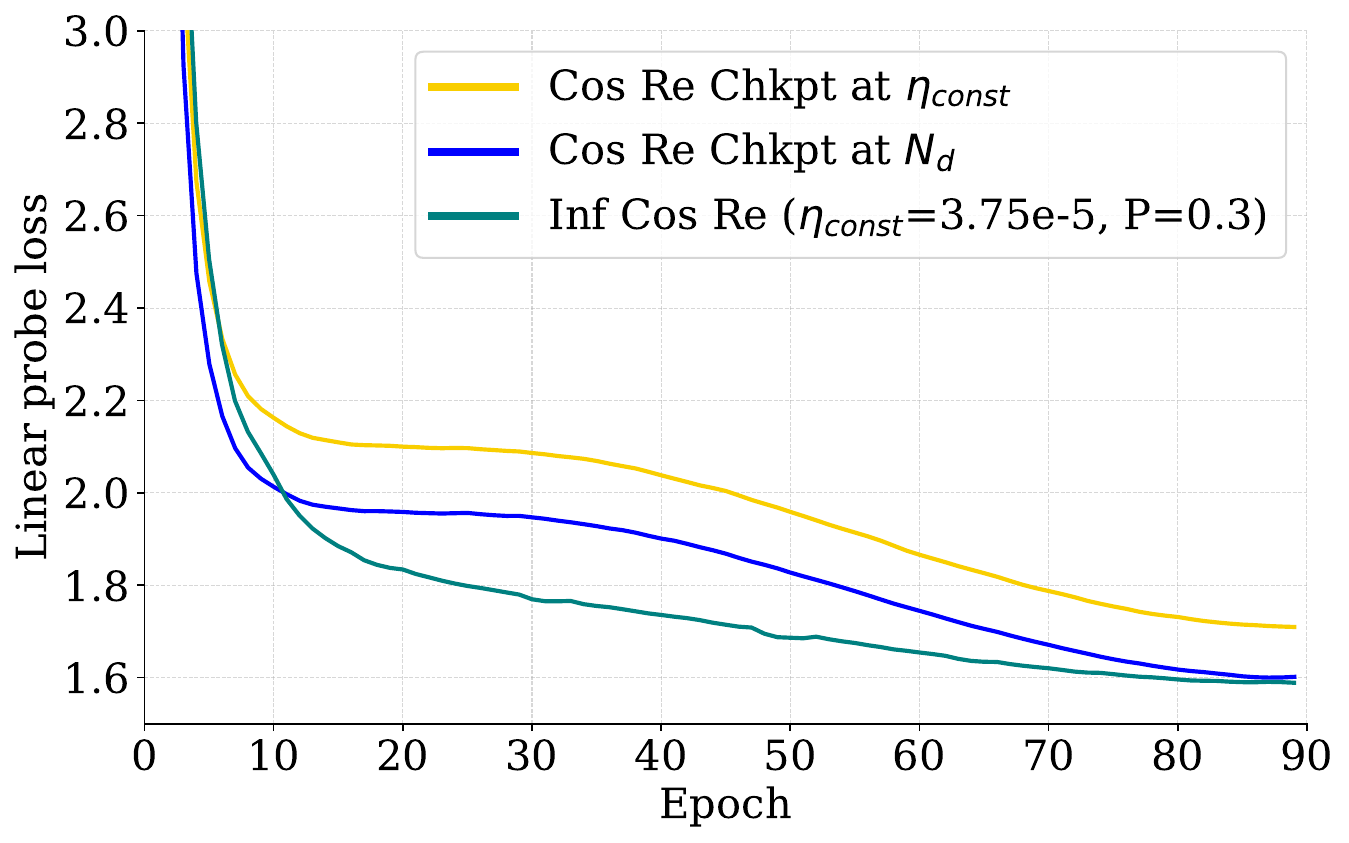}}
    
    \caption{\small Comparison of linear probe loss on Imagenet ($D_0$) with checkpoints obtained while training Places 2 ($D_1$). All experiments were conducted with replay. We compare different checkpoint selection strategies against the infinite scheduler. \textcolor{PineGreen}{Infinite Cosine learning rate schedule (green)} achieves consistently lower loss compared to cosine schedules restarted from either checkpoint at \textcolor{GreenYellow}{$\eta_{const}$ (yellow)} or \textcolor{blue}{$N_{d}$ (blue).}}
    \label{fig:rebuttal-checkpoint-ablation}
\end{figure}

To address the relative importance of the learning rate scheduling function versus the choice of checkpoint, we conducted an additional ablation study. \autoref{fig:rebuttal-checkpoint-ablation} shows the performance comparison of three different approaches:

\begin{enumerate}
    \item Cosine learning rate schedule with checkpoint from $\eta_{\text{const}}$ checkpoint (yellow line): This checkpoint is obtained at the point where the cosine schedule reaches the learning rate corresponding to the value of $\eta_{\text{const}}$ used in our infinite cosine schedule.
    \item Cosine learning rate schedule with checkpoint from $N_d$ checkpoint (blue line): This checkpoint is obtained at the time corresponding to when the annealing of the infinite schedule begins.
    \item Infinite Cosine learning rate schedule with $\eta_{\text{const}}=3.75\text{e-}5$ and $P=0.3$ (green line): Our best performing scheduler from main paper.
\end{enumerate}

As shown in \autoref{fig:rebuttal-checkpoint-ablation}, the Infinite Learning Rate schedule consistently achieves lower linear probe loss throughout training compared to both cosine schedule variants. The results demonstrate that this approach provides more effective representation learning compared to cosine schedule with early checkpointing.

Strategic checkpoint selection plays a critical role in preserving previously acquired knowledge. However, this flexibility is not inherently supported by the standard cosine schedule, which is typically designed to run uninterrupted until the end of training. In our comparison experiments, we manually extracted checkpoints at arbitrary points with high learning rates which is an approach that deviates from standard usage of cosine schedule. In contrast, the Infinite Learning Rate schedule naturally enables such flexibility through its constant learning rate phase, explicitly designed for this requirement. This built-in mechanism aids in knowledge retention across tasks, contributing to the consistently lower linear probe loss observed compared to other approaches.

\section{Additional Forgetting metrics}
\label{appen:forgetting_metrics}

We present a more thorough analysis of forgetting dynamics in our continual learning framework in \autoref{tab:Extended metrics}. While our main paper reports standard metrics which measured at the end of the training including Average Accuracy (AA), Forward Transfer (FWT), and Backward Transfer (BWT), this section extends our evaluation with additional metrics proposed by \citet{diaz2018don} to provide a more complete picture of the retention capabilities of our approach after every task.

\citet{diaz2018don} provides metrics to analyze the performance of the model at every timestep to incorporate the dynamic nature of CL. Hence we use an overall prefix for these metrics which are taken as an average after every time step. \textbf{Overall Accuracy} ($ A = \frac{\sum_{i \geq j}^{T} R_{i,j}}{\frac{T(T+1)}{2}}$) gives the average of accuracy on all tasks after every timestep. \textbf{Overall Backward transfer} ($BWT = \frac{\sum_{i=2}^{T} \sum_{j=1}^{i-1} \left( R_{i,j} - R_{j,i} \right)}{\frac{T(T-1)}{2}}$) measures the backward transfer after every timestep, while \textbf{ Overall Forward transfer} ($FWT = \frac{\sum_{i<j}^{T} R_{i,j}}{\frac{T(T-1)}{2}}$) measures the forward transfer after every time step. \textbf{Overall remembering } ($REM=100-|min(BWT,0)|$) quantifies the amount of knowledge remembered by the model across all time steps. Finally, we calculate the CL score to get a weighted average of all the metrics $CL_{score} = \sum_{i=1}^{\#c} w_i c_i$, where $\#c$ denotes the total number of metrics used. We use an equal weighting and report the equal weighted average. We additionally show the performance of a model which is pre-trained on the combination of all datasets, then linear probed on each dataset and obtained accuracy. We report the overall accuracy of it across all tasks as a reference in the table.

\begin{table}[]
\centering
\begin{tabular}{@{}lccccc@{}}
\toprule
\multicolumn{1}{c}{\textbf{Strategy}} & \multicolumn{1}{l}{\textbf{Overall Acc}} & \multicolumn{1}{l}{\textbf{Overall REM}} & \multicolumn{1}{l}{\textbf{Overall BWT}} & \multicolumn{1}{l}{\textbf{Overall FWT}} & \multicolumn{1}{l}{\textbf{CL\_score}} \\ \midrule
Cosine + ER (5\%)                     & 44.76                                    & 97.11                                    & -2.89                                    & 12.85                                    & 37.96                                  \\
Infinite cosine + ER (5\%)            & \textbf{50.56}                                    & \textbf{99.54}                                    & \textbf{-0.46 }                                   & 12.78                                    & \textbf{40.61}                                  \\
Cosine                                & 43.61                                    & 84.61                                    & -15.39                                   & 12.80                                    & 31.41                                  \\
Infinite cosine                       & 44.36                                    & 86.98                                    & -13.02                                   & \textbf{13.08}                                    & 32.85                                  \\ \midrule
Full baseline                         & 53.40                                   & -                                        & -                                        & -                                        & -                                      \\ \bottomrule
\end{tabular}
\caption{Comparison of forgetting metrics (as defined in \citet{diaz2018don}) across various methods. Infinite cosine with replay shows competitive results with lower forgetting, approaching the performance of the Full baseline.}
\label{tab:Extended metrics}
\end{table}

\section{Practical Guide for Hyperparameter Selection in Infinite Cosine Scheduling}

Selecting learning rate hyperparameters for continual pre-training can be computationally expensive due to the scale of training. This challenge applies to both repeated cosine and infinite cosine schedules and is not unique to the latter. However, when control over the initial pre-training phase is available, recent advances in hyperparameter-transfer techniques~\cite{yang2021tuning} can help reduce tuning costs substantially.

While hyperparameter stability is not the primary focus of this work, we provide practical guidance based on consistent trends observed in our experiments. The following empirically grounded rule-of-thumb can serve as a starting point for configuring the infinite cosine schedule effectively:

\begin{itemize}
    \item \textbf{Step 1: Selecting $\eta_{max}$} - Choose a maximum learning rate ($\eta_{max}$) that yields stable validation loss on the initial domain. This follows standard large-scale pre-training practice. If a tuned cosine schedule is already available from prior work, its $\eta_{max}$ serves as a strong candidate.
    
    \item \textbf{Step 2: Setting $\eta_{const}$} - Define the constant learning rate for Inf-Cos as the midpoint between $\eta_{max}$ and $\eta_{min}$. For example, with $\eta_{max} = 3e-4$ and $\eta_{min} = 3e-5$, a suitable baseline is \(\eta_{\text{const}} \approx 1.65 \times 10^{-4}\).
\end{itemize}

We further find the following trends to be consistent along with takeaways from \cite{gele2024scaling}

\begin{itemize}
    \item \textbf{Without replay:} When no replay buffer is available, a relatively lower $\eta_{const}$ is preferred to better preserve previously acquired knowledge as shown in \autoref{fig:code_val_loss_without_replay} and \autoref{fig:german_code_val_loss_without_replay}.

   \item \textbf{With replay:} When replay is available, a relatively higher $\eta_{const}$ can be utilized to improve adaptability without severely impacting retention — as shown in \autoref{fig:code_val_loss_with_replay}.
\end{itemize}

These insights allow practitioners to configure Infinite Schedules without exhaustive search, and makes it a more flexible and robust scheduler as compared to repeated cosine.

\section{Additional Experiments: Continual Learning with Medical Imaging}

To evaluate the robustness of our MAE continual pretraining approach under extreme domain shifts, we conduct additional experiments incorporating a fourth task that introduces a substantial distribution change. Specifically, we transition from natural image datasets to specialized medical imaging data, presenting a challenging scenario for knowledge retention and adaptation.

In this extended evaluation, we implement a continual learning scenario with an unlimited buffer capacity while constraining the replay mechanism through batch composition. During training on the fourth task, we enforce a balanced sampling strategy where 50\% of each batch consists of samples from the current task (MRI data) and the remaining 50\% comprises samples uniformly drawn from the combined buffer containing all previous datasets (ImageNet, Places2, and FireRisk).

The medical imaging dataset consists of multi-planar brain MRI scans. The dataset includes four distinct classes:``Empty" (background tissue),``Whole Tumor", ``Tumor Core", and ``Enhancing Tumor". For evaluation, we compute the average prediction accuracy across all relevant binary classification from the multi-class labels.

Table~\ref{tab:table-mri} presents the linear probe accuracies across all four tasks, comparing our proposed infinite learning rate schedule against the standard cosine annealing baseline. The infinite learning rate schedule consistently outperforms cosine annealing in mitigating catastrophic forgetting, maintaining higher accuracies on ImageNet (57.20\% vs 54.11\%), Places2 (47.16\% vs 46.46\%), and FireRisk (60.56\% vs 60.32\%) after completing all tasks. These results validate the effectiveness of our approach in continual learning scenarios involving substantial domain shifts.

\begin{table}[]
\centering
\begin{tabular}{@{}cccccccrr@{}}
\toprule
\textbf{Task completed} & \multicolumn{2}{c}{\textbf{Acc on Imagenet}} & \multicolumn{2}{c}{\textbf{Acc on Places}}                    & \multicolumn{2}{c}{\textbf{Acc on FireRisk}}                  & \multicolumn{2}{c}{\textbf{Acc on MRI}}                       \\ \midrule
\multicolumn{1}{l}{}    & Cosine                & Infinite             & Cosine                        & Infinite                      & Cosine                        & Infinite                      & \multicolumn{1}{c}{Cosine}    & \multicolumn{1}{c}{Infinite}  \\
After Imagenet          & \textbf{60.72}        & 59.84                & \cellcolor[HTML]{CCCCCC}46.11 & \cellcolor[HTML]{CCCCCC}45.70 & \cellcolor[HTML]{CCCCCC}59.80 & \cellcolor[HTML]{CCCCCC}59.89 & \cellcolor[HTML]{CCCCCC}85.49 & \cellcolor[HTML]{CCCCCC}85.46 \\
After Places            & 58.89                 & \textbf{60.35}       & \textbf{48.71}                & 48.37                         & \cellcolor[HTML]{CCCCCC}60.24 & \cellcolor[HTML]{CCCCCC}60.73 & \cellcolor[HTML]{CCCCCC}85.50 & \cellcolor[HTML]{CCCCCC}85.46 \\
After Firerisk          & 54.35                 & \textbf{57.13}       & 45.98                         & \textbf{47.23}                & 61.06                         & \textbf{61.29}                & \cellcolor[HTML]{CCCCCC}85.48 & \cellcolor[HTML]{CCCCCC}85.47 \\
After MRI               & 54.11                 & \textbf{57.20}       & 46.46                         & \textbf{47.16}                & 60.32                         & \textbf{60.56}                & \textbf{85.93}                & 85.92                         \\ \bottomrule
\end{tabular}
\caption{Linear probe accuracy comparison across four continual learning tasks using 50\% buffer replay. The infinite learning rate schedule consistently outperforms cosine annealing in mitigating catastrophic forgetting across diverse domains. Gray cells indicate forward transfer evaluation (tasks not yet encountered).}
\label{tab:table-mri}
\end{table}

\end{document}